\documentclass{article}
\usepackage{iclr2027_conference,times}

\usepackage{amsmath,amsfonts,bm}

\def\eqref#1{equation~\ref{#1}}

\def\1{\bm{1}}

\DeclareMathAlphabet{\mathsfit}{\encodingdefault}{\sfdefault}{m}{sl}
\SetMathAlphabet{\mathsfit}{bold}{\encodingdefault}{\sfdefault}{bx}{n}

\usepackage{booktabs}
\usepackage{array}
\usepackage{multirow}
\usepackage{graphicx}
\usepackage{xcolor}
\usepackage{hyperref}
\usepackage{url}
\usepackage{listings}

\usepackage{enumitem}
\usepackage{caption}
\newcommand{\modelname}{ERR-VLM}
\newcommand{\benchname}{ERR-Bench}
\newcommand{\envname}{ERR-Env}

\title{Replan for Error Recovery: Training Embodied VLMs through Verifiable Interaction with Action-Grouped Turn-GSPO}

\author{Zhuoran Wang$^{1,2}$,
  Keqing Fan$^{3}$, 
  Yang Li$^{1,2}$, 
  Jinbo Su$^{1,2}$, 
  Sijia Luo$^{1,2}$, 
  Ke Wang$^{1,2}$, \\
  \textbf{Yuanzheng Tan$^{4}$ , Junkun Peng$^{4}$ , Jing Zhang$^{1,5*}$} \\
  $^{1}$School of Information, Renmin University of China \\
  $^{2}$Key Laboratory of Data Engineering and Knowledge Engineering, Beijing, China \\
  $^{3}$Beihang University \\
  $^{4}$Zphyture \\
  $^{5}$Engineering Research Center of Database and Business Intelligence, Beijing, China
  }

\iclrfinalcopy

\makeatletter
\g@addto@macro\@thanks{%
  \footnotetext[1]{Corresponding author.}%
}
\makeatother

\begin{document}
\maketitle

\begin{abstract}
Embodied VLMs can execute increasingly complex manipulation tasks, yet execution errors can leave the environment in states that require replanning for recovery.
Existing recovery methods either rely on inference-time mechanisms with fixed policies or learn from failure examples and reference corrections rather than verified action outcomes.
To enable recovery learning through verifiable interaction, we introduce the Embodied Replanning for Recovery Environment (ERR-Env), which provides feedback on task progress and execution errors.
We further construct \benchname{}, a benchmark with held-out tasks spanning nominal, intermediate, and error states for evaluating recovery capability and execution reliability.
We then propose Action-Grouped Turn-GSPO (AGT-GSPO), an online reinforcement learning method combining turn-level rewards for task progress and error states with action-grouped advantage estimation, and use it to train \modelname{}, a compact 2B embodied VLM.
Evaluation on \benchname{} reveals a mismatch between task progress and execution reliability across diverse VLMs. AGT-GSPO improves \modelname{} by 10.25 percentage points in Strict Success Rate for error-free completion and 22.95 percentage points in Recovery Rate, while ablations and zero-shot transfer further validate the effectiveness of our method.

\end{abstract}

\section{Introduction}
\label{sec:introduction}

Vision-language models (VLMs) are increasingly used as high-level 
decision makers for embodied agents, enabling multi-step manipulation through visual observations, language goals, and semantic action interfaces \citep{pmlr-v202-driess23a, robobrain2026, hyembodied2026}. 
During execution, errors, limited visibility, or unexpected scene conditions can leave the task in a state for which the original plan is no longer appropriate.
Reliable task completion therefore requires the model to recognize these deviations and replan from the resulting state to recover and continue execution.

To support recovery from execution errors, inference-time approaches equip VLMs with prompting strategies, external monitors, or specialized mechanisms to detect errors and trigger replanning \citep{yu2024failureprompt, zhou2025codeasmonitor, ahmad2025failurehandling}.
But with the policy fixed, recovery is limited to existing model capabilities.
Recovery-oriented post-training instead uses error cases and reference corrections to improve error recognition and corrective action selection \citep{robofac2025, guardian2026, embodiedr15}.
However, reference-based supervision does not optimize the policy against verified outcomes of recovery actions, leaving a gap between imitating corrections and resolving execution errors.
This motivates training through environment interaction, where action outcomes are verified for error correction and task progress to improve recovery decisions.

In this paper, we aim to train embodied VLMs to recover from execution errors through verifiable interaction and evaluate their recovery capability and execution reliability. Interaction-based recovery learning requires explicit feedback on errors and task progress, which existing embodied benchmarks typically lack \citep{ma2025embodiedbench,voli2026volo}. 
To address this gap, we develop the \textbf{E}mbodied \textbf{R}eplanning for \textbf{R}ecovery Environment (\envname{}), an executable and verifiable interaction environment built on RoboTwin 2.0 \citep{mu2025robotwin}. 
\envname{} uses task-specific checkers to verify task progress, completion, execution errors, and visibility, supporting recovery-oriented data generation, online reinforcement learning, and evaluation. 
To assess recovery capability and execution reliability beyond task progress alone, we construct \benchname{} based on \envname{}, a controlled benchmark with held-out tasks spanning nominal, intermediate, and error states.

To learn recovery from verifiable interaction in \envname{}, verified action outcomes need to be translated into signals that reinforce effective corrections and discourage error-inducing actions. 
However, a trajectory-level signal cannot assign distinct feedback to error-inducing and corrective actions.
Therefore, we propose Action-Grouped Turn-GSPO (AGT-GSPO), an online reinforcement learning method that uses turn-level rewards capturing task progress and error states after each executed action. 
AGT-GSPO further estimates advantages within action groups, avoiding comparisons across action types. 
Using AGT-GSPO, we train \modelname{}, a compact 2B embodied VLM that learns to recognize execution errors and select corrective actions through replanning.

Evaluation of diverse VLMs on \benchname{} shows that high task progress does not guarantee reliable execution. Against this backdrop, training \modelname{} with AGT-GSPO improves both task completion and recovery, raising Strict Success Rate for error-free task completion by 10.25 percentage points and Recovery Rate by 22.95 percentage points over its supervised initialization. Ablations and zero-shot transfer to external simulators and a physical robot further validate our approach.

In summary, our principal contributions comprise:
\begin{itemize}[leftmargin=*]

\item We introduce \envname{}, an executable and verifiable interaction environment with feedback on task progress and execution errors, supporting recovery data generation, online RL, and evaluation. We further construct \benchname{} to evaluate recovery capability and execution reliability.

\item We propose AGT-GSPO, an online RL method that learns from verifiable interaction by combining turn-level rewards for task progress and error states with action-grouped advantage estimation. Using AGT-GSPO, we train \modelname{}, a compact 2B embodied VLM for error recovery.

\item Evaluation on \benchname{} across diverse VLMs reveals a clear mismatch between task progress and reliable execution, while \modelname{}, trained with AGT-GSPO, substantially improves both recovery capability and execution reliability.

\end{itemize}

\section{Related Work}
\label{sec:related}

\noindent\textbf{Interaction Environments and Benchmarks.}
Existing embodied environments and benchmarks support long-horizon interaction and evaluation \citep{shridhar2020alfred,padmakumar2022teach,srivastava2022behavior,ma2025embodiedbench,mu2025robotwin}, but largely focus on task completion and general capabilities rather than explicit verification of execution errors and recovery progress. Failure-aware benchmarks further evaluate error detection and correction \citep{song2026faer,voli2026volo}, yet lack verifiable interaction feedback for recovery learning and controlled evaluation. In contrast, \envname{} provides explicit error and progress feedback for recovery learning, while \benchname{} evaluates recovery capability and execution reliability under controlled error states.

\noindent\textbf{Error Recognition and Recovery.}
Inference-time methods use prompting, external monitors, or dedicated mechanisms to detect errors and trigger replanning \citep{yu2024failureprompt,zhou2025codeasmonitor,ahmad2025failurehandling}, but leave the policy fixed, limiting recovery to existing model capabilities. Training-based methods improve error awareness through error detection, reasoning, and verification \citep{duan2025aha,guardian2026}, but do not directly optimize corrective actions. More recent work supervises error correction and recovery planning \citep{robofac2025,embodiedr15}, but mainly relies on constructed error cases and reference corrections rather than verified action outcomes. We instead learn recovery through executable interaction with error and progress feedback.

\noindent\textbf{Online RL for Multi-turn Interaction.}
For general LLM agents, GRPO and GSPO use group-relative objectives for policy optimization \citep{deepseekmath2024,gspo2025}, while GiGPO extends this paradigm to multi-turn interaction with finer-grained credit assignment at both the episode and interaction-step levels \citep{gigpo2025}. In embodied settings, prior work applies offline GRPO without environment interaction \citep{reinforcedreasoning2025}, while online methods learn from grouped trajectories with completion-level rewards \citep{embodiedplanner2025}, lacking fine-grained turn-level information. VAGEN and ERA further introduce online turn-level rewards with GAE-based credit assignment \citep{vagen2025,era2025}, but rely on value-function estimation. In contrast, we combine verifiable turn-level error and progress feedback with group-relative advantage estimation within action groups for critic-free recovery optimization.

\begin{figure*}[!t]
    \centering
    \includegraphics[width=\textwidth]{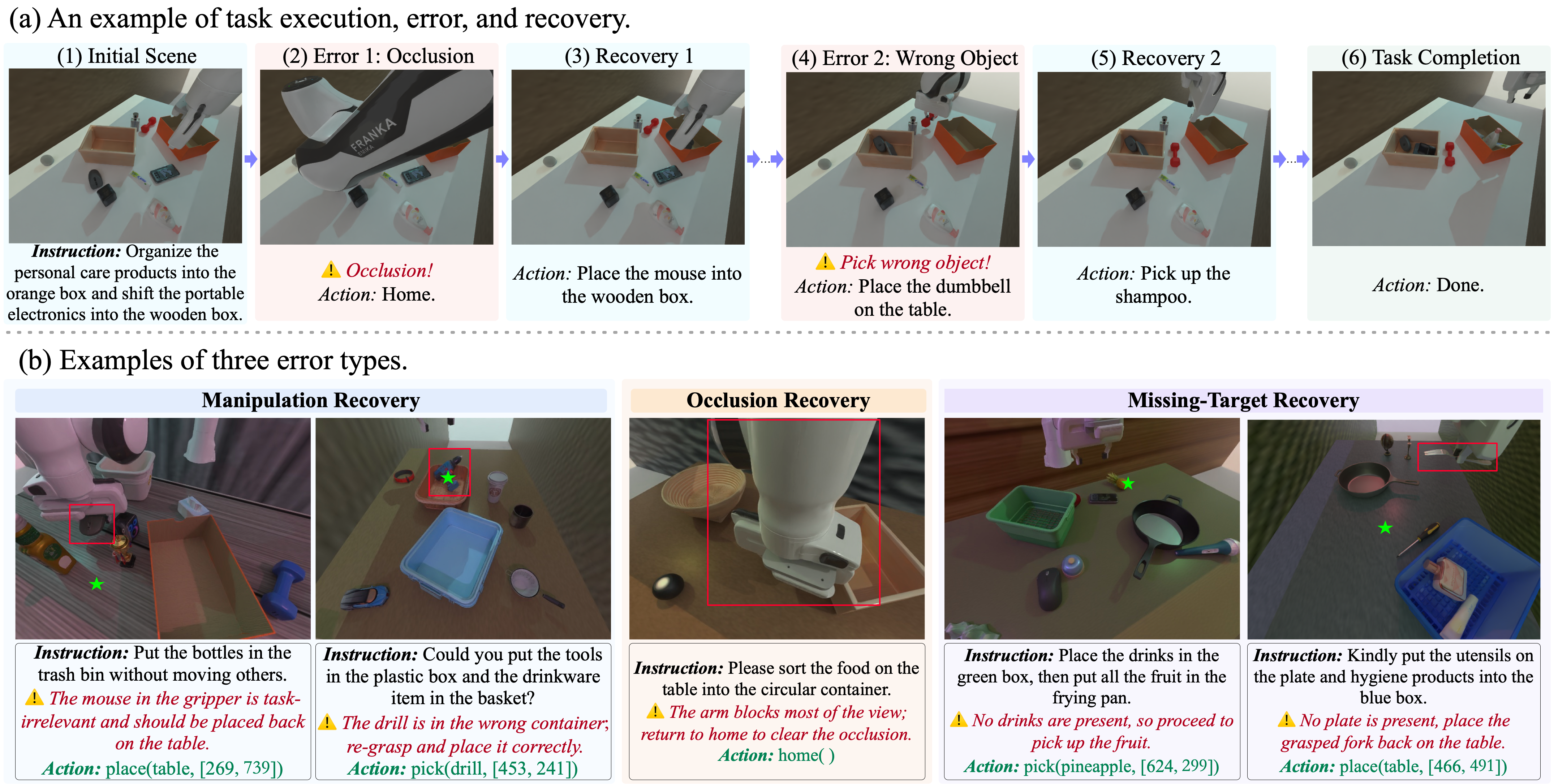}
    \caption{Recovery scenarios in \envname{}. (a) A trajectory with errors and corrective replanning. (b) Examples of Manipulation Recovery, Occlusion Recovery, and Missing-Target Recovery.}
    \label{fig:error-scenarios}
    \vspace{-8pt}
\end{figure*}

\section{Method}
\label{sec:method}

\noindent\textbf{Overview.}
We train embodied VLMs for recovery through verifiable interaction with three components:
First, \envname{} supports reproducible interaction from saved states and verifies progress and errors (\S \ref{sec:error-aware-env}).
Second, \benchname{} evaluates recovery and execution reliability on held-out nominal, intermediate, and error states (\S \ref{sec:err-bench}).
Third, AGT-GSPO learns recovery in \envname{} with turn-level rewards, action-grouped advantage estimation, and semantic advantage shaping (\S \ref{sec:turn-rl}).

\subsection{ERR-Env}
\label{sec:error-aware-env}

We develop \envname{} for recovery learning through interaction with verifiable feedback.
This requires reproducible task states and verifiable feedback on error and progress.
Built on the infrastructure and assets of RoboTwin 2.0 \citep{mu2025robotwin}, \envname{} provides restorable snapshots, controlled error scenarios, and task-specific checkers for progress, errors, completion, and visibility.

\noindent\textbf{Task Definition.}
Each task in \envname{} is represented as $\mathcal{T} = (\mathcal{X}, s_0, L, \mathcal{A}, \mathcal{C})$, where $\mathcal{X}$ is the executable environment interface, $s_0$ is the initial snapshot, $L$ is the task instruction, $\mathcal{A}$ is the action trajectory, and $\mathcal{C}$ is the task-specific checker. 
The interface $\mathcal{X}$ executes actions in the environment, including \texttt{pick}, \texttt{place}, \texttt{home}, and \texttt{done}. The \texttt{pick} and \texttt{place} actions operate on simulator objects, \texttt{home} returns the robot arm to its home pose, and \texttt{done} terminates execution. A task trajectory is defined as $\mathcal{A}=(a_0,\ldots,a_{n-1})$, with each $a_i$ specifying one of these actions.

A snapshot $s_i$ stores the complete robot and scene state, from which the corresponding visual observation can be rendered and the task state can be restored. Given $s_i$ and $a_i$, the interface executes the action and produces the next snapshot $s_{i+1} = \mathcal{X}(s_i,a_i)$.

\noindent\textbf{Controlled Error States.} For each recovery task, we construct an action trajectory containing an erroneous action and its corresponding recovery action:
\begin{equation}
\mathcal{A}^{\mathrm{rec}} = (a_0,\ldots,a_k,a_{\mathrm{error}},a_{\mathrm{recovery}},a_{k+1},\ldots,a_{n-1}).
\label{eq:recovery-sequence}
\end{equation}
Executing $a_{\mathrm{error}}$ moves the task into an error state, which is saved as an error snapshot. Executing $a_{\mathrm{recovery}}$ then produces a recovery state from which task execution can continue.

We consider three recovery categories: Manipulation Recovery (MR), where an incorrect object relation must be corrected; Occlusion Recovery (OR), where the robot arm occludes the workspace and the model must re-observe the scene; and Missing-Target Recovery (MTR), where a required object or container is unavailable and the model must replan from the resulting state. Figure~\ref{fig:error-scenarios}(a) illustrates the execution--error--recovery trajectory, while Figure~\ref{fig:error-scenarios}(b) shows representative examples of the three recovery categories.

\noindent\textbf{Task-Specific Checker.}
Each task includes an executable checker $\mathcal{C}$ implemented through a predefined interface. Given a snapshot $s$, the checker evaluates task-relevant object relations, manipulation states, and object visibility in the rendered observation, and returns 
\begin{equation}
\mathcal{C}(s) = \bigl(\mathrm{success}(s), \mathrm{progress}(s), \mathrm{penalty}(s), \mathrm{visibility}(s)\bigr).
\end{equation}
Here, $\mathrm{success}$ indicates task completion; $\mathrm{progress}$ measures correct advancement, including partial completion; $\mathrm{penalty}$ measures erroneous state changes, such as misplaced objects or grasping task-irrelevant objects; and $\mathrm{visibility}$ counts workspace objects visible in the rendered observation. Checker outputs are used only for training rewards and evaluation, not exposed to the policy.

\noindent\textbf{Task Generation.} We generate executable recovery tasks in four stages:
\begin{itemize}
\item An asset pool provides candidate assets for task construction, from which the LLM selects the objects, distractors, and containers used in each task.
\item Given the selected assets, the LLM generates a task instruction and a language-level semantic-action sequence, containing the designated error and recovery actions.
\item Using a predefined task-construction interface, the LLM converts the selected assets, instruction, and semantic-action sequence into an executable program that defines scene initialization, action trajectory execution, and the task-specific checker.
\item We execute the generated program to verify task completion and discard tasks that fail to execute successfully.
\end{itemize}

\subsection{ERR-Bench} \label{sec:err-bench}
We build \benchname{} on \envname{} as a data resource for recovery-oriented training and evaluation. It provides trajectories for SFT, snapshots for online RL, and held-out tasks spanning nominal, intermediate, and error states for controlled evaluation.

\noindent\textbf{Data Generation.}
For each validated task, we generate multiple instances with randomized object poses, backgrounds, viewpoints, and robot configurations. Executing the task trajectory produces visual observations and intermediate snapshots. For Occlusion Recovery, visibility is checked after each action; once the target is occluded, the state is saved as an error snapshot, followed by a \texttt{home} action to generate the recovery snapshot.

\noindent\textbf{Training Data.}
We partition validated tasks into training and held-out sets. Training instances are further divided between SFT and online RL. SFT data consist of task instructions, visual observations, and action trajectories, augmented with instruction paraphrases and VLM-generated reasoning grounded by simulator-derived boxes. For online RL, nominal, intermediate and error snapshots from the remaining instances are used to initialize interactive rollouts.

\noindent\textbf{Evaluation Suite.}
The evaluation suite of \benchname{} is constructed exclusively from held-out tasks unseen during SFT and online RL, with novel action trajectories, instruction realizations, and asset instances. It contains 1,503 snapshots from distinct held-out tasks, including initial, intermediate, and controlled error states for full-task execution, continued execution, and recovery. The three state categories are equally represented, with error states evenly split across the three error types.

\subsection{Action-Grouped Turn-GSPO}
\label{sec:turn-rl}

We train an embodied VLM to recover from execution errors by learning from verified feedback that distinguishes effective actions from invalid or non-improving ones while accounting for action-type differences. We initialize the policy with recovery-aware SFT and propose AGT-GSPO, combining turn-level rewards, action-grouped advantage estimation, and semantic advantage shaping.

\begin{figure*}[t]
    \centering  \includegraphics[width=\textwidth]{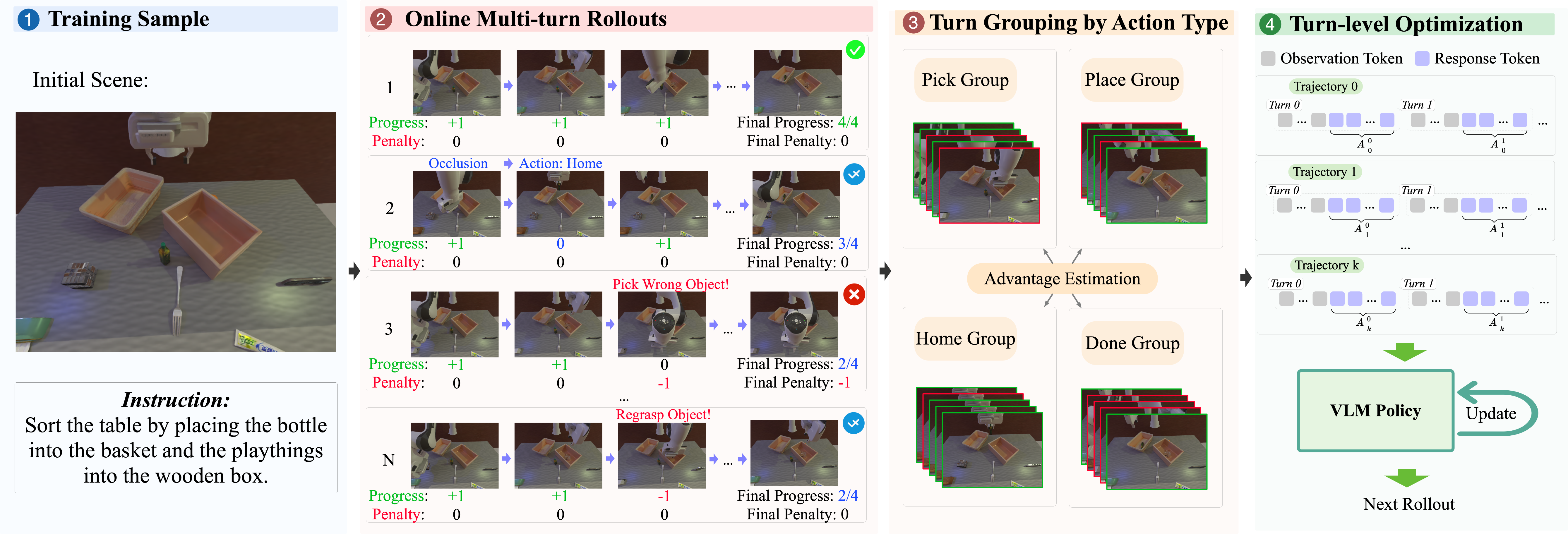}
    \caption{Training process of AGT-GSPO. The policy interacts with \envname{} over multiple turns and receives turn-level feedback on task progress and error states. Turns are grouped by action type for relative advantage estimation, enabling credit assignment among comparable recovery decisions.}
    \label{fig:training-overview}
    \vspace{-8pt}
\end{figure*}

\noindent\textbf{Recovery-Aware SFT.} We initialize \modelname{} from Qwen3-VL-2B-Thinking \citep{qwen3vl2025} and use SFT as a cold start for online RL. Each sample contains the task instruction, current observation, and interaction history for multi-turn cases. The model generates grounded reasoning and a tool call, with \texttt{pick} and \texttt{place} additionally predicting a 2D image-space point that is converted to a 3D target by the environment interface. From 20,435 self-collected episodes, we construct 65,000 training samples: about one-third are full multi-turn trajectories, while the remainder are individual turns from nominal, intermediate, and recovery states.

Starting from the SFT policy, we apply AGT-GSPO to learn recovery through interaction in \envname{}.  AGT-GSPO constructs turn-level rewards from grounding quality, feedback on task progress and error states, and rollout outcomes, then estimates and shapes action-grouped advantages before turn-level GSPO optimization. Figure~\ref{fig:training-overview} summarizes the online rollout and optimization procedure.

\noindent\textbf{Grounding Reward.} The grounding reward $r_t^{\mathrm{grd}}$ evaluates grounding quality using simulator-derived object boxes. It combines IoU-based localization accuracy with coverage of visible manipulable objects and containers in the robot workspace to preserve object localization and spatial reasoning. Details of box matching and reward formulation are provided in Appendix~\ref{app:reward}.

\noindent\textbf{Action Reward.}
The action reward $r_t^{\mathrm{act}}$ provides turn-level feedback directly from the task-specific checker. We define a checker-based state score as
\begin{equation}
q(s)
=
\mathrm{progress}(s)
-
\mathrm{penalty}(s),
\label{eq:state-score}
\end{equation}
where $\mathrm{progress}$ measures correct task advancement and $\mathrm{penalty}$ measures erroneous task changes. The action reward is
\begin{equation}
r_t^{\mathrm{act}}
=
\begin{cases}
r_t^{\mathrm{exec}} + q(s_{t+1})-q(s_t),
& a_t\in\{\texttt{pick},\texttt{place}\},\\
r_t^{\mathrm{exec}} + \mathrm{visibility}(s_{t+1})-\mathrm{visibility}(s_t),
& a_t=\texttt{home},\\
r_t^{\mathrm{exec}} + \mathrm{success}(s_{t+1}),
& a_t=\texttt{done}.
\end{cases}
\label{eq:action-reward}
\end{equation}
Here, $r_t^{\mathrm{exec}}$ evaluates whether the predicted tool call is valid and executable, including correct formatting, valid grounding, and satisfaction of action preconditions. The remaining terms use checker outputs to measure task progress and error changes, visibility recovery, and successful termination.

\noindent\textbf{Outcome Reward.}
The outcome reward $r_t^{\mathrm{out}}$ provides rollout-level feedback on overall task performance. For a rollout $\tau$, we define
\begin{equation}
R^{\mathrm{out}}(\tau)
=
R^{\mathrm{success}}(\tau)
+R^{\mathrm{step}}(\tau)
+R^{\mathrm{done}}(\tau),
\label{eq:outcome-reward}
\end{equation}
where $R^{\mathrm{success}}$ measures final task progress, $R^{\mathrm{step}}$ measures execution efficiency, and $R^{\mathrm{done}}$ measures termination correctness. We assign $r_t^{\mathrm{out}}=R^{\mathrm{out}}(\tau)$ to each turn in the rollout. The detailed definitions and weighting of these terms are provided in Appendix~\ref{app:reward}.

The immediate reward for turn $t$ is
\begin{equation}
r_t = \lambda_{\mathrm{grd}} r_t^{\mathrm{grd}}
    + \lambda_{\mathrm{act}} r_t^{\mathrm{act}}
    + \lambda_{\mathrm{out}} r_t^{\mathrm{out}}.
\label{eq:turn-reward}
\end{equation}

\noindent\textbf{Action-Grouped Advantage Estimation.} 
Different action types have distinct reward semantics and scales, making normalization across actions potentially biased by action-specific reward distributions. We therefore normalize each turn within the same action type. For \texttt{pick} and \texttt{place}, we use same-type turns within the sample's rollout group. Since \texttt{home} and \texttt{done} are less frequent, we instead use same-type turns across the current update batch for a larger comparison group:
\begin{equation}
A_t^{\mathrm{raw}}
=
\frac{r_t-\mu_{\mathcal G_t}}
{\sigma_{\mathcal G_t}+\delta},
\qquad
\mathcal G_t=
\begin{cases}
\{u\in\mathcal O_i \mid a_u=a_t\},
& a_t\in\{\texttt{pick},\texttt{place}\},\\
\{u\in\mathcal B \mid a_u=a_t\},
& a_t\in\{\texttt{home},\texttt{done}\},
\end{cases}
\label{eq:raw-adv}
\end{equation}
where $\mathcal O_i$ is the rollout group for sample $i$, $\mathcal B$ is the current update batch, and $\mu_{\mathcal G_t}$ and $\sigma_{\mathcal G_t}$ denote the mean and standard deviation of the rewards of the turns in $\mathcal G_t$, respectively. The constant $\delta>0$ is included for numerical stability. This intentionally removes cross-action reward-scale differences, making the advantage a within-action relative signal.

\noindent\textbf{Semantic Advantage Shaping.}
Favorable rollout-level outcomes can assign positive relative advantages to invalid or non-improving turns, while low within-group variance can weaken the directional signal for clearly improving turns. We therefore further shape the raw advantage for \texttt{pick} and \texttt{place} using execution validity and task-state change. Let $\Delta q_t=q(s_{t+1})-q(s_t)$. We define
\begin{equation}
\widetilde{A}_t =
\begin{cases}
\min\!\left(A_t^{\mathrm{raw}},\mu_{\mathrm{inv}}\right),
& r_t^{\mathrm{exec}} < 0,\\
\min\!\left(A_t^{\mathrm{raw}},\mu_{\mathrm{deg}}\right),
& r_t^{\mathrm{exec}} \geq 0,\ \Delta q_t \leq 0,\\
\max\!\left(A_t^{\mathrm{raw}},\mu_{\mathrm{imp}}\right),
& r_t^{\mathrm{exec}} \geq 0,\ \Delta q_t > 0.
\end{cases}
\label{eq:semantic-advantage-shaping}
\end{equation}
Here, $\mu_{\mathrm{inv}}<0$, $\mu_{\mathrm{deg}}<0$, and $\mu_{\mathrm{imp}}>0$ are small signed thresholds. For \texttt{home} and \texttt{done}, we directly use $\widetilde{A}_t=A_t^{\mathrm{raw}}$.

\noindent\textbf{Turn-Level GSPO Optimization.}
Let $I_t$ denote the current rendered observation, $H_t$ the preceding interaction history, and $y_t$ the turn-$t$ response containing all reasoning tokens and the serialized action. We optimize $y_t$ as a single sequence with shaped advantage $\widetilde{A}_t$. Given current policy $\pi_\theta$ and rollout policy $\pi_{\mathrm{old}}$, turn-level GSPO uses the sequence-level importance ratio
\begin{equation}
\rho_t(\theta)=\left(\frac{\pi_\theta(y_t\mid I_t,L,H_t)}{\pi_{\mathrm{old}}(y_t\mid I_t,L,H_t)}\right)^{1/|y_t|},
\label{eq:gspo-ratio}
\end{equation}
and optimizes the clipped objective, where $\epsilon>0$ is the clip radius and $\beta$ weights a KL regularizer:
\begin{equation}
\mathcal{L}_{\mathrm{GSPO}}(\theta)=
\mathbb{E}_{\tau,t}\left[\min\left(\rho_t(\theta)\widetilde{A}_t,
\operatorname{clip}(\rho_t(\theta),1-\epsilon,1+\epsilon)\widetilde{A}_t\right)\right]
-\beta\,\mathcal{L}_{\mathrm{KL}}.
\label{eq:turn-gspo}
\end{equation}

\section{Experiments}
\label{sec:experiments}

To evaluate recovery capability, execution reliability, and transferability, we organize our experiments into three parts. First, we compare execution and recovery performance against representative VLMs on \benchname{} in Section~\ref{sec:main-results}. Second, we analyze AGT-GSPO components through ablations in Section~\ref{sec:ablation}. Third, we evaluate transfer on long-horizon tasks in two external simulators and assess both task completion and recovery with a physical robot in Section~\ref{sec:zero-shot}.

\noindent\textbf{Setup.}
We train \modelname{} for 4,500 SFT steps at a learning rate of $5\times10^{-6}$, followed by 200 RL steps at $2\times10^{-6}$. The main evaluation and ablation studies use the held-out \benchname{} evaluation set of 1,503 snapshots. Zero-shot transfer is evaluated on RMBench and RoboVoLo, as well as in a real-world setting with an AIRBOT robot arm.

\subsection{Main Results on \benchname}
\label{sec:main-results}

We compare the compact SFT policy and the full \modelname{} policy against closed-source models (Qwen3.7-Plus~\citep{qwen2026qwen37plus}, GPT-5.6 Sol~\citep{openai2026gpt56sol}, and Qwen3.8-Max~\citep{qwen2026qwen38max}), open-source VLMs (Qwen3-VL-235B-A22B~\citep{qwen3vl2025}, Qwen3-VL-2B-Thinking~\citep{qwen3vl2025}, and InternVL3.5-8B~\citep{internvl2025}), and embodied VLMs (RoboBrain2.5-8B~\citep{robobrain2026}, Hy-Embodied-VLM-1.0~\citep{hyembodied2026}, MiMo-Embodied-7B~\citep{hao2025mimoembodied}, and the error-aware Embodied-R1.5~\citep{embodiedr15}). All models use the same system prompt, tool definitions, OpenAI-compatible tool-call format, and a 500-token thinking budget. Table~\ref{tab:main-results} reports aggregate execution and recovery results.

We report Success Rate (SR), the fraction of episodes completing all subtasks; Progress (PG), the fraction of completed subtasks; No-Penalty Rate (NPR), the fraction of episodes without irrelevant grasps or object misplacement; and Strict Success Rate (SSR), which requires both task completion and error-free termination. For recovery evaluation, we consider three error types: Manipulation Recovery (MR), Occlusion Recovery (OR), and Missing-Target Recovery (MTR). For each type, Recovery Rate measures whether the induced error is corrected, while Recovery SSR further requires strict completion of the remaining task. Total denotes the average across the three error types.

\begin{table*}[t]
\caption{Main results on \benchname. We evaluate all tasks with four metrics: Success Rate (SR), Progress (PG), No-Penalty Rate (NPR), and Strict Success Rate (SSR). Recovery Rate and Recovery SSR are reported for Manipulation Recovery (MR), Occlusion Recovery (OR), Missing-Target Recovery (MTR), and their average (Total). All entries are percentages.}
\centering
{\fontsize{8}{9.5}\selectfont
\setlength{\tabcolsep}{0pt}
\begin{tabular}{@{}>{\raggedright\arraybackslash}p{0.20\textwidth}@{\hspace{1pt}\vrule width 0.2pt\hspace{1pt}}*{4}{>{\centering\arraybackslash}p{0.0653\textwidth}}@{\hspace{1pt}\vrule width 0.2pt\hspace{1pt}}*{4}{>{\centering\arraybackslash}p{0.0653\textwidth}}@{\hspace{1pt}\vrule width 0.2pt\hspace{1pt}}*{4}{>{\centering\arraybackslash}p{0.0653\textwidth}}@{}}
\toprule
\multirow{2}{*}{Model} & \multicolumn{4}{c@{\hspace{1pt}\vrule width 0.2pt\hspace{1pt}}}{All Tasks} & \multicolumn{4}{c@{\hspace{1pt}\vrule width 0.2pt\hspace{1pt}}}{Recovery Rate} & \multicolumn{4}{c}{Recovery SSR} \\
 & SR & PG & NPR & SSR & MR & OR & MTR & Total & MR & OR & MTR & Total \\
\midrule
\multicolumn{13}{l}{\textit{Closed-source models}} \\
GPT-5.6 Sol & \underline{40.98} & \underline{76.81} & 51.97 & 15.44 & \underline{38.92} & 53.29 & 55.09 & 49.10 & 14.97 & 11.98 & 8.38 & 11.78 \\
Qwen3.8-Max & \textbf{55.16} & \textbf{82.66} & 53.36 & \textbf{27.21} & 30.54 & \underline{85.63} & 56.89 & \underline{57.69} & \underline{25.75} & \textbf{43.71} & \underline{10.18} & \textbf{26.55} \\
Qwen3.7-Plus & 18.16 & 52.58 & 54.89 & 5.39 & 23.35 & 37.13 & 51.50 & 37.33 & 9.58 & 7.78 & 3.59 & 6.98 \\
\midrule
\multicolumn{13}{l}{\textit{Open-source models}} \\
Qwen3-VL-235B-A22B & 26.55 & 68.90 & 40.85 & 10.25 & 32.34 & 37.13 & 37.72 & 35.73 & 14.97 & 16.17 & 1.80 & 10.98 \\
InternVL3.5-8B & 9.45 & 48.94 & 46.77 & 3.33 & 21.56 & 0.00 & \underline{65.27} & 28.94 & 1.20 & 0.00 & 3.59 & 1.60 \\
Qwen3-VL-2B (Base) & 8.05 & 40.26 & 66.34 & 2.06 & 16.17 & 0.00 & 64.67 & 26.95 & 1.20 & 0.00 & 0.00 & 0.40 \\
\midrule
\multicolumn{13}{l}{\textit{Embodied VLMs}} \\
RoboBrain2.5-8B & 12.78 & 49.21 & 42.64 & 4.06 & 17.37 & 23.95 & 53.29 & 31.54 & 4.79 & 0.60 & 1.80 & 2.40 \\
Hy-Embodied-VLM-1.0 & 6.92 & 40.38 & \underline{67.80} & 4.05 & 15.57 & 0.00 & 60.48 & 25.35 & 3.59 & 0.00 & 1.20 & 1.60 \\
MiMo-Embodied-7B & 7.79 & 38.80 & \textbf{69.33} & 0.86 & 16.76 & 0.00 & 52.69 & 23.15 & 1.20 & 0.00 & 0.00 & 0.40 \\
Embodied-R1.5 & 10.85 & 54.86 & 41.58 & 3.39 & 17.37 & 4.19 & 51.50 & 24.35 & 4.19 & 0.00 & 0.00 & 1.40 \\
\midrule
\multicolumn{13}{l}{\textit{Ours}} \\
\modelname{} (SFT, 2B) & 18.96 & 62.36 & 53.09 & 10.64 & 26.95 & 40.72 & 59.88 & 42.52 & 13.17 & 4.19 & 8.38 & 8.58 \\
\modelname{} (Full, 2B) & 34.14 & 71.08 & 62.41 & \underline{20.89} & \textbf{40.12} & \textbf{89.22} & \textbf{67.07} & \textbf{65.47} & \textbf{26.35} & \underline{24.55} & \textbf{15.57} & \underline{22.16} \\
\bottomrule
\end{tabular}
}
\label{tab:main-results}
\vspace{-12pt}
\end{table*}

\noindent\textbf{Task progress and recovery do not necessarily imply reliable completion.}
Qwen3.8-Max achieves the highest SR of 55.16 and PG of 82.66, yet its SSR is 27.21, while an NPR of 53.36 indicates that high progress can still involve irrelevant grasps or incorrect placements. Conversely, MiMo-Embodied-7B obtains a high NPR of 69.33 but only 38.80 PG and 0.86 SSR, showing that a model with limited task-completion ability may incur few penalties simply because it makes little effective progress. A similar pattern appears in recovery, where all models exhibit a severe drop from Recovery Rate to Recovery SSR; even the highest Recovery SSR is only 26.55. This gap shows that correcting an immediate failure does not necessarily lead to reliable task completion afterward, highlighting the importance of evaluating both recovery success and reliable task completion.

\noindent\textbf{Recovery-oriented training improves both task completion and recovery capability.}
Compared with the SFT policy, the full \modelname{} improves all overall and recovery metrics, increasing SR from 18.96 to 34.14, SSR from 10.64 to 20.89, Recovery Rate from 42.52 to 65.47, and Recovery SSR from 8.58 to 22.16. Across recovery settings, stronger large models generally show better error-recovery performance than smaller open-source and embodied models, reflecting a clear capability gap in recovery. This gap is especially pronounced in Occlusion Recovery: Qwen3.8-Max reaches a Recovery Rate of 85.63, while several smaller models remain near zero, revealing their limited ability to re-observe and recover. Recovery-oriented training markedly improves the recovery performance of our 2B model across different failure types, showing that targeted training can endow compact models with strong recovery capabilities that do not naturally emerge at this scale.

\noindent\textbf{Overall.}
These results highlight a gap between task progress and reliable completion in \benchname{}. Meanwhile, \modelname{} improves task and recovery performance, showing that recovery-oriented training strengthens general execution and error recovery in compact embodied models.

\subsection{Ablation}
\label{sec:ablation}

We ablate five components of the recovery-oriented training pipeline: recovery data, reward design, group construction, training component, and optimization strategy. Starting from the same SFT initialization, we remove recovery data, ablate individual reward signals, vary how interaction samples are grouped for optimization, remove semantic advantage shaping, and compare turn-level and trajectory-level RL objectives. Table~\ref{tab:ablations} summarizes their effects on both task execution and recovery.

\begin{table*}[t]
\captionsetup{
    justification=raggedright,
    singlelinecheck=false
}
\caption{Ablation results on \benchname. Metrics follow Table~\ref{tab:main-results}. All entries are percentages.}
\centering
{\fontsize{8}{9.5}\selectfont
\setlength{\tabcolsep}{0pt}
\begin{tabular}{@{}>{\raggedright\arraybackslash}p{0.22\textwidth}@{\hspace{1pt}\vrule width 0.2pt\hspace{1pt}}*{4}{>{\centering\arraybackslash}p{0.0636\textwidth}}@{\hspace{1pt}\vrule width 0.2pt\hspace{1pt}}*{4}{>{\centering\arraybackslash}p{0.0636\textwidth}}@{\hspace{1pt}\vrule width 0.2pt\hspace{1pt}}*{4}{>{\centering\arraybackslash}p{0.0636\textwidth}}@{}}
\toprule
\multirow{2}{*}{Method} & \multicolumn{4}{c@{\hspace{1pt}\vrule width 0.2pt\hspace{1pt}}}{All Tasks} & \multicolumn{4}{c@{\hspace{1pt}\vrule width 0.2pt\hspace{1pt}}}{Recovery Rate} & \multicolumn{4}{c}{Recovery SSR} \\
 & SR & PG & NPR & SSR & MR & OR & MTR & Total & MR & OR & MTR & Total \\
\midrule
\multicolumn{13}{l}{\textit{SFT and full RL}} \\
\modelname{} (SFT) & 18.96 & 62.36 & 53.09 & 10.64 & 26.95 & 40.72 & 59.88 & 42.52 & 13.17 & 4.19 & 8.38 & 8.58 \\
\modelname{} (Full) & 34.14 & 71.08 & \underline{62.41} & \textbf{20.89} & 40.12 & \textbf{89.22} & \underline{67.07} & \textbf{65.47} & \textbf{26.35} & \underline{24.55} & \textbf{15.57} & \textbf{22.16} \\
\midrule
\multicolumn{13}{l}{\textit{Recovery-data Ablations}} \\
w/o Recovery Data in RL & 25.03 & 69.14 & 52.77 & 14.17 & 28.14 & 28.74 & 60.48 & 39.12 & 13.17 & 7.78 & 13.77 & 11.57 \\
\midrule
\multicolumn{13}{l}{\textit{Reward Ablations}} \\
w/o Grounding Reward & 32.54 & 72.35 & 56.16 & 16.31 & \underline{43.11} & 80.24 & 56.89 & 60.08 & \underline{23.35} & 22.16 & \underline{14.37} & \underline{19.96} \\
w/o Action Reward & \textbf{36.53} & 74.21 & 43.32 & 10.05 & 38.92 & 87.43 & 46.71 & 57.69 & 17.96 & 12.57 & 7.19 & 12.57 \\
w/o Penalty Feedback & 32.01 & 72.45 & 43.25 & 2.80 & 35.93 & 83.83 & 51.50 & 57.09 & 10.18 & 7.78 & 5.39 & 7.78 \\
w/o Outcome Reward & 30.55 & 71.33 & 49.57 & 11.78 & 31.14 & 80.24 & 65.27 & 58.88 & 18.56 & 11.98 & 12.57 & 14.37 \\
\midrule
\multicolumn{13}{l}{\textit{Group Ablations}} \\
Batch (All) & \underline{35.93} & \underline{77.28} & 58.89 & \underline{19.83} & 38.92 & 83.83 & 66.47 & \underline{63.07} & 16.17 & \textbf{28.14} & 10.18 & 18.16 \\
Sample (All) & 31.81 & 75.66 & 50.70 & 9.25 & 33.53 & 88.02 & 56.89 & 59.48 & 11.98 & 8.98 & 6.59 & 9.18 \\
Batch (Action) & 31.61 & \textbf{77.72} & 47.91 & 11.11 & 34.73 & \underline{88.62} & 46.71 & 56.69 & 11.98 & 4.19 & 10.18 & 8.78 \\
\midrule
\multicolumn{13}{l}{\textit{RL Component Ablations}} \\
w/o Advantage Shaping & 27.21 & 68.14 & 51.43 & 10.64 & 33.53 & 71.86 & 53.29 & 52.89 & 16.76 & 17.36 & 10.79 & 14.97 \\
\midrule
\multicolumn{13}{l}{\textit{Algorithm Ablations}} \\
Turn-GRPO & 31.81 & 74.89 & 53.89 & 8.85 & \textbf{45.51} & 83.83 & 48.50 & 59.28 & 13.77 & 10.18 & 9.58 & 11.18 \\
Trajectory-GSPO & 10.92 & 51.10 & \textbf{65.41} & 8.05 & 17.37 & 56.29 & \textbf{67.66} & 47.11 & 7.78 & 1.20 & 7.78 & 5.59 \\
Trajectory-GRPO & 13.91 & 58.10 & 60.02 & 9.45 & 23.95 & 59.88 & 65.27 & 49.70 & 6.59 & 1.20 & 7.19 & 4.99 \\
\bottomrule
\end{tabular}
}
\label{tab:ablations}
\vspace{-12pt}
\end{table*}

\noindent\textbf{Recovery data and reward design provide complementary supervision.}
Removing recovery data reduces Recovery Rate from 65.47 to 39.12 while PG remains similar, showing that nominal interaction alone is insufficient for learning recovery. Reward ablations separate progress from action correctness. Without the action reward, PG stays high but NPR and SSR drop due to weaker action-level supervision. Removing penalty feedback is more harmful: erroneous actions that still advance the task can receive positive progress signals without explicit penalties, causing SSR to fall to 2.80. This shows that misleading feedback on incorrect actions is worse than simply removing an action-level signal. Grounding and outcome rewards provide additional supervision on state-action alignment and task completion, further stabilizing policy learning and improving execution quality.

\noindent\textbf{Group construction benefits from meaningful comparisons.}
Our full method groups the same action type within each sample for pick and place, preserving both action-level comparability and shared task context. Sample (All) keeps the sample context but mixes different action types, while Batch (Action) preserves action type but compares turns from different samples; both weaken strict execution and recovery, showing that either source of mismatch can reduce the quality of relative advantage estimation. Batch (All) relaxes both constraints and performs comparatively well, likely because the larger comparison set yields a more stable estimate. However, its Recovery SSR drops from 22.16 to 18.16, while the overall Recovery Rate also decreases from 65.47 to 63.07, indicating that broader grouping weakens both strict execution performance and coarse recovery signals.

\noindent\textbf{Turn-level optimization benefits from semantic advantage shaping and GSPO.}
Removing semantic advantage shaping lowers Recovery Rate from 65.47 to 52.89 and Recovery SSR from 22.16 to 14.97. The small correction suppresses misleading positive advantages on invalid or non-improving turns and strengthens weak signals on improving turns. Trajectory-level GSPO and GRPO perform substantially worse, indicating that trajectory-level feedback is too coarse for multi-turn recovery. Turn-GRPO achieves higher PG but lower SSR and Recovery SSR than our full ERR-VLM. Unlike GRPO's token-wise reweighting, GSPO applies sequence-level weighting and clipping, yielding more coherent updates for each action turn and better execution reliability.

\noindent\textbf{Overall.}
The ablations show that the full method benefits from complementary choices across data, rewards, grouping, and optimization. Each component contributes to more reliable task execution and recovery, while removing any of them degrades either recovery quality or strict execution.

\subsection{Zero-Shot Transfer}
\label{sec:zero-shot}

\begin{table*}[t]
\captionsetup{
    justification=raggedright,
    singlelinecheck=false
}
\caption{Results on RMBench and RoboVoLo. Each task reports Progress in percentages.}
\centering
{\scriptsize
\setlength{\tabcolsep}{1pt}
\begin{tabular*}{\textwidth}{@{\extracolsep{\fill}}l*{3}{c}|*{4}{c}@{}}
\toprule
\multirow{2}{*}{Model} & \multicolumn{3}{c|}{RMBench} & \multicolumn{4}{c}{RoboVoLo} \\
& Observe and Pick Up & Cover Blocks & Place Block Mat & Memory & Common Sense & World Knowledge & Complex References \\
\midrule
RoboBrain2.5-8B & 65.0 & 33.3 & 38.3 & 62.5 & 95.5 & 20.0 & 45.0 \\
ERR-VLM (SFT) & 50.0 & 55.0 & 46.7 & 37.0 & 52.1 & 15.0 & 38.3 \\
ERR-VLM (Full) & 90.0 & 56.7 & 55.0 & 72.8 & 96.3 & 25.0 & 59.2 \\
\bottomrule
\end{tabular*}
}
\label{tab:transfer-sim}
\vspace{-5pt}
\end{table*}

\begin{table*}[t]
\centering
\begin{minipage}[c]{0.69\textwidth}
    \leftskip=0pt\rightskip=0pt\parfillskip=0pt plus 1fil
    \refstepcounter{table}\label{tab:transfer-real}
    \noindent Table \thetable: Real-world experiment results. PG and NPR are reported for each task, and Active Re-observation Rate (ARR) is reported overall. ARR measures the probability of invoking \texttt{home} after an occlusion to restore visual observation. All entries are percentages. \\ \par
    \centering
    {\fontsize{6.5}{7.5}\selectfont
    \setlength{\tabcolsep}{2.5pt}
    \resizebox{\linewidth}{!}{%
    \begin{tabular}{l*{11}{c}}
    \toprule
    \multirow{2}{*}{Model} & \multicolumn{2}{c}{2 Targets} & \multicolumn{2}{c}{3 Targets} & \multicolumn{2}{c}{4 Targets} & \multicolumn{2}{c}{MR} & \multicolumn{2}{c}{MTR} & \multicolumn{1}{c}{Overall} \\
    & PG & NPR & PG & NPR & PG & NPR & PG & NPR & PG & NPR & ARR \\
    \midrule
    RoboBrain2.5-8B & 47.5 & 20.0 & 48.3 & 70.0 & 25.7 & 20.0 & 55.0 & 10.0 & 72.5 & 40.0 & 8.9 \\
    ERR-VLM (SFT) & 42.5 & 100.0 & 55.0 & 100.0 & 21.4 & 70.0 & 45.0 & 40.0 & 45.0 & 80.0 & 0.0 \\
    ERR-VLM (Full) & 70.0 & 100.0 & 73.3 & 100.0 & 40.0 & 80.0 & 55.0 & 60.0 & 70.0 & 90.0 & 78.7 \\
    \bottomrule
    \end{tabular}
    }
    }
\end{minipage}\hfill
\begin{minipage}[c]{0.29\textwidth}
    \centering
    \includegraphics[width=\linewidth]{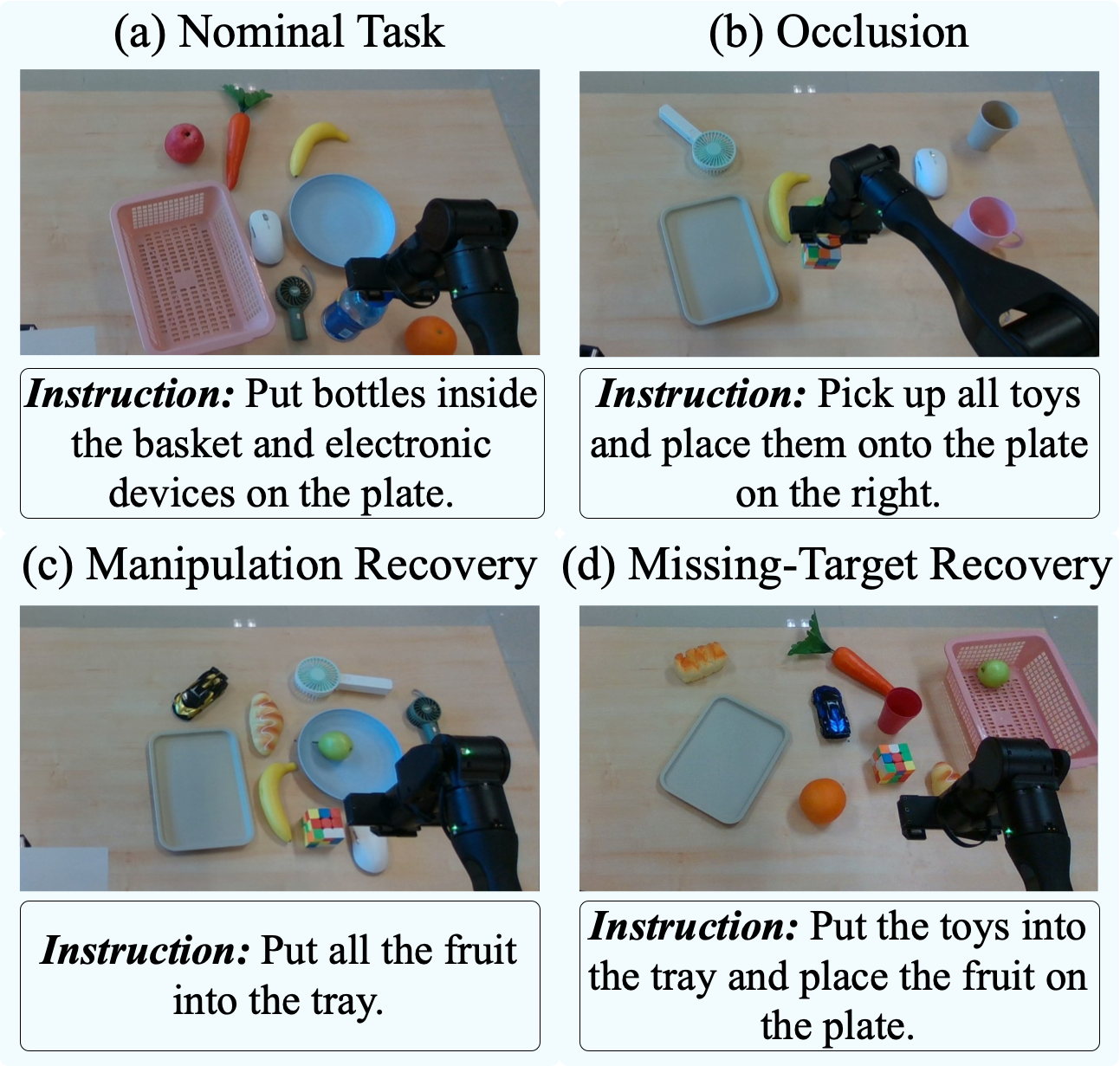}
    \par\vspace{0pt}
    \leftskip=0pt\rightskip=0pt\parfillskip=0pt plus 1fil
    \refstepcounter{figure}\label{fig:real-task-exp}
    \noindent Figure \thefigure: Real-world scenes.\par
\end{minipage}
\vspace{-12pt}
\end{table*}

We further evaluate planning and recovery transfer beyond the training environment. RoboBrain2.5-8B, our SFT policy, and the full \modelname{} policy are tested zero-shot on RMBench, RoboVoLo \citep{chen2026rmbench,voli2026volo}, and a real-world robot. We evaluate three RMBench pick-and-place tasks, four RoboVoLo task types, and five real-world tasks covering long-horizon manipulation, Manipulation Recovery, and Missing-Target Recovery. Each model is evaluated for 10 trials per task under the same action space, without additional demonstrations, parameter updates, or reward tuning. Results are reported in Table~\ref{tab:transfer-sim} and Table~\ref{tab:transfer-real}, with representative real-world examples shown in Figure~\ref{fig:real-task-exp}.

\noindent\textbf{Simulation transfer.}
Across RMBench and RoboVoLo, the full \modelname{} consistently outperforms its SFT initialization across all evaluated categories and achieves performance comparable to RoboBrain2.5-8B. Gains span long-horizon manipulation, memory, reasoning, and recovery, indicating that RL improves capabilities that transfer beyond the training distribution.

\noindent\textbf{Real-world transfer.}
The gains from SFT to the full \modelname{} policy also hold in real-world evaluation. The full policy improves PG across all long-horizon tasks and achieves higher NPR on both MR and MTR, indicating more reliable recovery with fewer incorrect grasps. We further report Active Re-observation Rate (ARR), defined as the probability of invoking \texttt{home} after occlusion to restore visual observation. ARR increases from 0.0\% for the SFT policy to 78.7\% after RL training, showing that the policy learns to reacquire visual information before continuing execution.

\noindent\textbf{Overall.}
Across simulation and real-world settings, RL consistently improves over SFT, enhancing execution, recovery, and active re-observation. These results demonstrate that our training effectively learns behaviors that transfer zero-shot beyond the training environment.

\section{Conclusion}
\label{sec:conclusion}
In this paper, we train embodied VLMs to recover from execution errors through verifiable interaction and evaluate their recovery capability and execution reliability. \envname{} provides executable interaction with feedback on task progress and execution errors, enabling recovery data generation, online RL, and evaluation. Built on \envname{}, \benchname{} enables controlled evaluation across nominal, intermediate, and error states. AGT-GSPO learns from verified action outcomes through turn-level rewards and action-grouped advantage estimation. Experiments show that task progress alone does not ensure reliable execution, while recovery-oriented training improves both task completion and recovery and transfers to external simulators and real-world tasks. Overall, verifiable interaction provides an effective basis for learning and evaluating recovery in embodied VLMs.

\noindent\textbf{Limitations.}
Our study adopts a deliberately controlled scope to investigate embodied error recovery. The current setting is limited in three aspects: we consider a compact high-level action space and a limited range of error types; training is conducted in a single simulation environment with a checker tailored to pick-and-place tasks; and, due to computational constraints, we primarily train a 2B VLM rather than systematically studying model scaling. These choices allow us to isolate recovery-related behaviors and evaluate the proposed learning framework under controlled conditions, while leaving broader settings beyond the current scope. Future work will extend the framework to broader tasks, environments, recovery scenarios, and model scales.

\section*{Reproducibility Statement}

We provide sufficient detail to reproduce the proposed recovery-oriented training and evaluation pipeline. Section~\ref{sec:method} describes the core methodology, including \envname{}, \benchname{}, the checker-based reward design, and Action-Grouped Turn-GSPO. Building on this, Appendix~\ref{app:env-construct} details the construction of executable tasks, the data generation and annotation procedures, the semantic action execution interface, and the serialization of multi-turn tool-call trajectories. Appendix~\ref{app:reward} specifies the reward terms and advantage-shaping thresholds, while Appendix~\ref{app:training-configuration} reports the complete SFT and RL training setup, including model initialization, optimization hyperparameters, hardware, and training duration. Section~\ref{sec:experiments} presents the main experiments, ablations, and zero-shot transfer evaluations, with the evaluation metrics defined in Appendix~\ref{app:metrics} and further experimental details provided in Appendix~\ref{app:addition-exp}. We additionally include anonymized implementation code and scripts in the supplementary material to facilitate reproduction of the main experiments.
\section*{AI Use Statement}
Generative AI tools were used in several aspects of this work, including data generation and data annotation as part of the research pipeline, as described in the corresponding sections of the paper. They were also used to assist with software development, including code generation, debugging, and the organization of experimental scripts, as well as for language editing and translation during manuscript preparation. All AI-assisted outputs, including generated data, annotations, code, and manuscript text, were reviewed and verified by the authors. The authors were responsible for the design of the research and experimental procedures, the validation and quality control of the generated data and annotations, the correctness of the implementation and experimental results, and the interpretation and conclusions presented in the paper, and take full responsibility for the final content of this work.

\bibliography{iclr2027_conference}
\bibliographystyle{iclr2027_conference}

\clearpage
\appendix
\section{Appendix}

\subsection{Executable Environment Construction}
\label{app:env-construct}

\subsubsection{Task Definition}
\label{app:task-def}

We formulate the task as sequential interaction between a policy and an executable robotic environment. Starting from a given environment snapshot, the policy observes the current scene and issues a sequence of semantic actions through an executable interface $\mathcal{X}$. The interface is responsible for restoring the environment state, grounding and validating the selected actions, executing the corresponding robot behaviors, and returning execution outcomes.

\noindent\textbf{Task Definition.}
Each generated task is represented as $\mathcal{T}=(\mathcal{X},s_0,L,\mathcal{A},\mathcal{C})$, where $\mathcal{X}$ is the executable environment interface, $s_0$ is the initial snapshot, $L$ is the instruction, $\mathcal{A}$ is the reference action sequence, and $\mathcal{C}$ is the task-specific checker. Executing $a_i$ from snapshot $s_i$ yields $s_{i+1}=\mathcal{X}(s_i,a_i)$; all generated intermediate states are saved as loadable snapshots.

\noindent\textbf{Executable Environment Interface.}
The executable environment interface $\mathcal{X}$ maps each semantic action---\texttt{pick}, \texttt{place}, \texttt{home}, or \texttt{done}---to an environment transition. It loads the current snapshot, grounds the policy's image-space point to a visible scene actor, checks action preconditions, executes the fixed motion primitive, and returns an ordinary execution result. The result is appended to the interaction history.

\noindent\textbf{$\bullet$~Environment Loading.}
For an execution starting from snapshot $s_i$, the executor restores the corresponding simulator state in full. This restoration reconstructs the scene background and camera viewpoint, reinstantiates all scene objects together with their recorded poses, and recovers the spatial and task-relevant relations among them. It also restores the robot state, including the arm configuration and gripper state. If the snapshot indicates that the gripper is already holding an object, the corresponding object is reattached to the gripper with its recorded relative pose so that execution resumes from a physically consistent state. During execution, the executor tracks whether the gripper currently holds an object and whether the episode has terminated; these state variables are used to enforce the preconditions of subsequent semantic actions.
\begin{lstlisting}
class ActionExecutor:
    def __init__(self, snapshot):
        self.env = load(snapshot)
        self.held_object = self.get_holding_object()
        self.task_finished = False
\end{lstlisting}

\noindent\textbf{$\bullet$~Action Dispatch.}
The executor reads the semantic action name and dispatches it to the corresponding primitive. Unknown action names yield an ordinary failed execution result rather than modifying the environment state.
\begin{lstlisting}
    def execute(self, action):
        name = action["action"]
        if name == "pick":
            return self.pick(action["label"], action["point_2d"])
        if name == "place":
            return self.place(action["label"], action["point_2d"])
        if name == "home":
            return self.home()
        if name == "done":
            return self.done()
        return {"success": False, "reason": "unknown_action"}
\end{lstlisting}

\noindent\textbf{$\bullet$~Pick and Place.}
For \texttt{pick} and \texttt{place}, $\mathcal{X}$ first checks the action-specific execution preconditions before invoking the low-level controller. A \texttt{pick} action is permitted only when the gripper is empty, whereas a \texttt{place} action requires the gripper to be holding an object. Once these preconditions are satisfied, the model-selected 2D image point is grounded to a scene actor using the current instance-segmentation observation. Specifically, $\mathcal{X}$ identifies the segmentation region containing the selected point and resolves it to the corresponding simulator actor. The grounded actor is then validated according to the action type; for example, a \texttt{pick} target must correspond to a graspable object, while points falling on non-graspable entities such as the table, robot, or other background structures are rejected. For \texttt{place}, the grounded actor must correspond to a valid placement target, such as a container or support surface. After a valid target actor is identified, the low-level executor directly queries its 3D pose from the simulator state and computes the corresponding robot end-effector pose required for grasping or placement. The executor then attempts the motion primitive multiple times to improve robustness. After each attempt, success is determined from the resulting physical state, such as whether the target object is actually held by the gripper after \texttt{pick}, or whether the held object satisfies the intended object--target relation after \texttt{place}. If an attempt fails, the simulator is restored to the state immediately preceding the action and another execution attempt is made. After the allowed attempts are exhausted or the action succeeds, $\mathcal{X}$ returns the final execution result and updates the held-object state only when the corresponding physical outcome has been verified.
\begin{lstlisting}
    def ground_object_at_point(self, point):
        rgb, depth, camera_pose = self.env.get_front_camera_observation()
        world_point = self.env.pixel_to_world(point, depth, camera_pose)
        return self.env.get_visible_actor_near(world_point)

    def pick(self, object_name, point):
        if self.held_object is not None:
            return {"success": False, "reason": "gripper_not_empty"}
        obj = self.ground_object_at_point(point)
        if obj is None:
            return {"success": False, "reason": "no_object_at_point"}
        if not obj.is_graspable:
            return {"success": False, "reason": "object_not_graspable"}
        if not self.env.is_reachable(obj):
            return {"success": False, "reason": "object_not_reachable"}
        grasp_pose = self.env.compute_grasp_pose(obj, point)
        success = self.env.pick_with_retry(obj, grasp_pose)
        if success and self.env.check_grasp(obj):
            self.held_object = obj
            return {"success": True, "reason": "object_grasped"}
        return {"success": False, "reason": "grasp_failed"}

    def place(self, container_name, point):
        if self.held_object is None:
            return {"success": False, "reason": "gripper_empty"}
        container = self.ground_object_at_point(point)
        if container is None:
            return {"success": False, "reason": "no_container_at_point"}
        if not container.is_container:
            return {"success": False, "reason": "invalid_placement_target"}
        if not self.env.is_reachable(container):
            return {"success": False, "reason": "container_not_reachable"}
        place_pose = self.env.compute_place_pose(
            object=self.held_object, container=container, point=point)
        if not self.env.is_collision_free(place_pose, self.held_object):
            return {"success": False, "reason": "placement_collision"}
        success = self.env.place_with_retry(
            target=self.held_object, container=container, place_pose=place_pose)
        if success and self.env.check_on(self.held_object, container):
            placed = self.held_object
            self.held_object = None
            return {"success": True, "reason": "object_placed",
                    "object": placed.name, "container": container.name}
        return {"success": False, "reason": "placement_failed"}
\end{lstlisting}

\noindent\textbf{$\bullet$~Home.}
The \texttt{home} action moves the robot to a predefined home configuration. For each robot embodiment, a robot-specific home pose is specified in advance such that the arm is positioned to minimize occlusion of the workspace and provide a clear observation for subsequent decisions. When \texttt{home} is executed, the robot returns to this predefined pose while preserving the current gripper state. In particular, if the gripper is holding an object, the object remains attached throughout the motion, and the held-object state is left unchanged.
\begin{lstlisting}
    def home(self):
        success = self.env.move_to_home_pose()
        if success:
            return {"success": True, "reason": "robot_at_home_pose"}
        return {"success": False, "reason": "home_motion_failed"}
\end{lstlisting}

\noindent\textbf{$\bullet$~Done.}
The \texttt{done} action terminates the current execution. Once invoked, no further actions are executed, and the current environment state is treated as the final state of the episode for subsequent task evaluation.
\begin{lstlisting}
    def done(self):
        self.execution_end()
        return {"success": True, "reason": "task_finished"}
\end{lstlisting}

\subsubsection{Task Generation}
\label{app:task-generation}

This section describes the generation of executable tasks. We introduce asset selection and action specification, followed by the task-construction interface and the resulting executable task code.

\noindent\textbf{Semantic Task Generation.}
This stage corresponds to the first two steps of the task-generation pipeline described in the main text. In the first step, an asset pool provides candidate objects, distractors, and containers, from which Qwen3.7-Max~\citep{qwen2026qwen37max} selects the assets required for each task. In the second step, conditioned on the selected assets, the model generates the task instruction together with a language-level semantic-action sequence that specifies the intended execution. For recovery tasks, we use a dedicated prompt for each error type so that the generated sequence follows the corresponding failure pattern and includes both the designated erroneous action and the subsequent recovery actions. A representative generated record is:
\begin{lstlisting}
{
  "scene": {"containers": [...], "objects": [...], "distractors": [...]},
  "task": "Place each object in its designated container.",
  "action": ["pick object_1; place container_1",
              "pick object_2; place container_2",
              "pick distractor; place table"],
  "error_type": "wrong_object",
  "instruction": "..."
}
\end{lstlisting}

\noindent\textbf{Task-Construction Interface.}
This stage corresponds to the third step of the task-generation pipeline described in the main text. The LLM is first provided with a fixed task-construction interface that defines the set of methods available for generating an executable task program. Through this interface, the model can instantiate scene actors, encode the reference semantic-action sequence, specify task completion, and define task-specific state checks. The interface restricts task generation to these predefined high-level operations and does not expose low-level robot control commands. The main interfaces available to the LLM are listed below.
\begin{lstlisting}
def add_actor(self, object_type, object_name):
    """Create an environment actor.
    Args: object_type is the asset type; object_name is the instance name.
    Returns: actor.
    """

def add_distractors(self, object_list):
    """Create distractor actors from a list of asset type names."""

def pick_and_place(self, object, container):
    """Pick an actor and place it in or on a target actor.
    Returns: success.
    """
  
def add_end(self):
    """Mark the end of the reference task sequence."""

def check_grasp(self, object):
    """Return whether the robot arm currently grasps an actor."""

def check_on(self, object, container):
    """Return whether an actor is in or on a target actor."""

def count_visible_objects(self, scene):
    """Return the number of visible objects in the given scene."""
\end{lstlisting}

\noindent\textbf{Executable task code.}
Using the task-construction interface described above, Qwen3.7-Max generates an executable task program conditioned on the selected assets, task instruction, reference semantic-action sequence, and error specification. Each generated program consists of three main components: environment initialization, execution of the reference semantic-action sequence, and a task-specific checker. The following representative Manipulation Recovery task places a Rubik's cube and knife into designated containers, while treating a candlestick and sampled distractors as irrelevant objects:
\begin{lstlisting}
class Task_1_toy_tableware_wrong_object_1(Task):

    def load_actors(self):
        # Containers.
        self.wooden_box = self.add_actor("wooden_box", "wooden_box")
        self.steamer = self.add_actor("steamer", "steamer")

        # Task-relevant objects.
        self.rubikscube = self.add_actor("rubikscube", "rubikscube")
        self.knife = self.add_actor("knife", "knife")

        # Designated irrelevant object for the wrong-object error.
        self.candlestick = self.add_actor("candlestick", "candlestick")

        # Additional distractors.
        self.distractors = self.add_distractors([
            "sand-clock", "soy-sauce", "bread", "boxdrink", "keyboard"])
        self.check_scene()

    def play_once(self):
        # Place each target in its designated container.
        success = self.pick_and_place(self.knife, self.steamer)
        if not success:
            return self.info

        success = self.pick_and_place(self.rubikscube, self.wooden_box)
        if not success:
            return self.info

        # Manipulate an irrelevant object, then return it to the table.
        success = self.pick_and_place(self.candlestick, self.table)
        if not success:
            return self.info

        self.add_end()
        return self.info

    def checker(self, success_mode=False, progress_mode=False,
                penalty_mode=False, visibility_mode=False):
        rubik_in_box = self.check_on(self.rubikscube, self.wooden_box)
        knife_in_steamer = self.check_on(self.knife, self.steamer)

        if success_mode:
            return rubik_in_box and knife_in_steamer

        if progress_mode:
            target_is_grasped = (self.check_grasp(self.rubikscube) 
            or self.check_grasp(self.knife))
            grasp_bonus = 0.5 if target_is_grasped else 0.0
            ground_truth_progress = 2.0
            progress = rubik_in_box + knife_in_steamer + grasp_bonus
            return progress, ground_truth_progress

        if visibility_mode:
            return self.count_visible_objects(self.scene)

        if penalty_mode:
            penalty = 0.0

            # Target objects in the wrong target container.
            rubik_in_steamer = self.check_on(self.rubikscube, self.steamer)
            knife_in_box = self.check_on(self.knife, self.wooden_box)
            if rubik_in_steamer:
                penalty += 1.0
            if knife_in_box:
                penalty += 1.0

            # Irrelevant objects must not be grasped or put in targets.
            irrelevant_objects = [self.candlestick] + self.distractors
            for obj in irrelevant_objects:
                object_is_grasped = self.check_grasp(obj)
                object_in_box = self.check_on(obj, self.wooden_box)
                object_in_steamer = self.check_on(obj, self.steamer)

                if object_is_grasped:
                    penalty += 0.5
                if object_in_box:
                    penalty += 1.0
                if object_in_steamer:
                    penalty += 1.0

            return penalty

        return rubik_in_box and knife_in_steamer
\end{lstlisting}
The environment initialization component instantiates the objects, robot embodiment, and camera configuration required by the task. Task-relevant objects, containers, and distractors are loaded from the asset pool, with their poses randomized during scene construction. The robot embodiment is selected from the available robot set, while the camera is sampled from a predefined set of viewpoints and further perturbed to introduce additional visual variation across samples. After scene construction, \texttt{check\_scene} verifies that the instantiated objects are visible from the selected camera view and that the resulting scene satisfies the basic requirements for task execution.

The execution component implements the reference semantic-action sequence generated in the preceding task-generation stage using the predefined task construction interface. It specifies the sequence of high-level manipulation operations required to realize the task, including the designated erroneous and recovery actions when applicable. This reference sequence is used only during data generation to execute the task, generate intermediate states and snapshots, and verify that the generated task is physically executable. It is never provided to the policy during rollout, reinforcement learning, or evaluation.

The checker defines task-specific signals for completion, progress, penalty, and visibility.
\begin{equation}
\mathcal{C}(s)=\bigl(\mathrm{success}(s),\mathrm{progress}(s),\mathrm{penalty}(s),\mathrm{visibility}(s)\bigr).
\end{equation}
The completion check determines whether the intended final task state has been achieved; the progress check measures partial task completion and is used as part of the reinforcement-learning reward; the penalty check detects undesired interactions, such as manipulating irrelevant objects or placing objects in incorrect containers; and the visibility check measures whether relevant scene entities remain observable. These checker outputs are used for data-generation validation, reward computation during reinforcement learning, and final evaluation, but are never exposed to the policy. To further ensure task correctness, generated tasks are validated both automatically, and a randomly sampled subset is additionally validated manually: the reference semantic-action sequence is executed to confirm that the task can be completed successfully and satisfies the checker, and a randomly sampled subset of generated task programs is additionally inspected by human annotators to verify that the scene configuration, action semantics, and checker logic are consistent with the intended task.

\subsubsection{Data Generation and Annotation}
\label{app:data-annotation}

This section describes the data generation and annotation process. We first execute and validate the generated task programs to collect observations and snapshots, and then construct SFT chain-of-thought (CoT) and grounding annotations from the validated trajectories.

\noindent\textbf{Generation Procedure.}
For each validated task, we generate multiple execution trajectories by repeatedly instantiating the task under independently randomized scene configurations. Across trajectories, we vary the background, object poses, object appearances, camera viewpoints, and robot embodiments, while preserving the underlying task semantics and reference semantic-action sequence. Each instantiated trajectory is then executed step by step in the simulator. At the initial state and after every executed action, we save the rendered observation together with the complete simulator snapshot, and record the corresponding semantic action that produces the transition. As a result, a single task yields multiple visually and physically diverse trajectories, each containing aligned observations, executable snapshots, and action records spanning the initial, intermediate, error, recovery, and terminal states encountered during execution.

\noindent\textbf{SFT CoT and Grounding Annotation.}
After trajectory validation, the multiple trajectories generated from the same task are paired with linguistically diverse instruction variants that preserve the underlying task semantics while varying the phrasing, reasoning style, and tone. We use Qwen3.7-Max to generate these instruction variants and to annotate each trajectory step with a concise chain-of-thought rationale describing the current state, task-relevant entities, and the next semantic action. To provide explicit visual grounding, instance masks obtained directly from the simulator are converted into image-space bounding boxes and associated with the entities referenced in the rationale. Each SFT sample therefore consists of an instruction variant, the corresponding rendered observation, a grounded rationale, and the valid serialized semantic action.

\subsubsection{Tool-Call Input and Multi-Turn Records}
\label{app:tool-data}

This section describes the tool-call interface and interaction records provided to the policy. We first specify the robot-action schema and system instruction, then describe how observations, reasoning, actions, and tool results are organized into single- and multi-turn records.

The policy receives the following tool schema. It exposes one function, \texttt{robot\_action}, whose required arguments are \texttt{action}, \texttt{label}, and \texttt{point\_2d}:
\begin{lstlisting}
{
  "type": "function",
  "function": {
    "name": "robot_action",
    "description": "Send exactly one robot manipulation command for the current observation.",
    "parameters": {
      "type": "object",
      "properties": {
        "action": {
          "type": "string",
          "enum": ["pick", "place", "home", "done"],
          "description": "The next robot command to send."
        },
        "label": {
          "type": "string",
          "description": "Object name for pick, target location/container name for place, or empty string for home and done."
        },
        "point_2d": {
          "type": "array",
          "description": "2D point [x, y] in 0-1000 image coordinates for pick/place, or [] for home and done.",
          "items": {"type": "integer"}
        }
      },
      "required": ["action", "label", "point_2d"]
    }
  }
}
\end{lstlisting}
The policy interacts with the environment through serialized tool calls, which are passed to the executable environment interface $\mathcal{X}$ for execution. For \texttt{pick} and \texttt{place}, \texttt{label} identifies the target object or destination, and \texttt{point\_2d} specifies an integer image coordinate $[x,y]$ in the range $[0,1000]$. For \texttt{home} and \texttt{done}, \texttt{label} is empty and \texttt{point\_2d} is an empty list. Given each tool call, $\mathcal{X}$ performs the corresponding semantic action and returns the resulting environment transition to the policy interaction history.

The system prompt used for all policies is:
\begin{lstlisting}
You are a robot manipulation agent.

Your job is to look at the current environment image and decide the single next robot action.

Rules:
- Use 'pick' when the robot should grasp an object.
- Use 'place' when the robot should place the currently manipulated object onto or into a target location.
- Use 'home' when the robot should return to its home position during occlusion.
- Use 'done' only when the task is already complete.

\end{lstlisting}

\noindent\textbf{Turn Samples.} 
Each multi-turn SFT trajectory is represented as an ordered \texttt{messages} sequence that records the interaction between the policy and the environment. At each decision step, the policy receives the current task instruction and visual observation and produces an assistant turn containing three aligned components: \texttt{reasoning\_content}, a textual action response, and a \texttt{robot\_action} tool call. The \texttt{reasoning\_content} summarizes the current scene, identifies task-relevant objects and containers, grounds referenced entities with image-space bounding boxes, and explains the choice of the next semantic action. The textual response provides a concise description of the selected action, while the tool call serializes the executable action and its target for execution through the environment interface $\mathcal{X}$. For \texttt{pick} and \texttt{place}, the tool call additionally specifies an integer image-space target point through \texttt{point\_2d}. The following example shows two consecutive assistant turns, illustrating the complete grounded reasoning, action response, and tool-call fields for a \texttt{pick} followed by a \texttt{place}.

\begin{lstlisting}
{"role": "assistant",
  "content": "Action: pick stapler",
  "reasoning_content": "The robot gripper is positioned above the center of the table, with a white plate([495.0, 589.0, 822.0, 1000.0]) containing a cucumber([559.0, 626.0, 640.0, 835.0]), baguette([660.0, 704.0, 962.0, 944.0]), and bottle on the right side. On the left side of the tabletop, a silver stapler([116.0, 415.0, 232.0, 511.0]) lies next to a blue toy car and a small glue stick([362.0, 457.0, 424.0, 522.0]), while a bell([0.0, 505.0, 91.0, 605.0]) sits further to the far left. The stapler is an office tool that belongs in the steamer([407.0, 315.0, 476.0, 480.0]) container rather than remaining on the table surface. Since the stapler is currently ungrasped and outside its designated destination, the correct next action is to pick up the stapler.",
  "tool_calls": [{
    "function": {
      "name": "robot_action",
      "arguments": {
        "action": "pick",
        "label": "stapler",
        "point_2d": [180, 457]}}}]}

{"role": "assistant",
  "content": "Action: place stapler on steamer",
  "reasoning_content": "The robot gripper holds a small stapler([85.0, 380.0, 211.0, 468.0]) directly above the perforated steamer([407.0, 284.0, 601.0, 484.0]) basket situated in the center of the table. To the left, a blue toy car and a red strawberry toy rest on the surface, while a white plate([495.0, 589.0, 822.0, 1000.0]) containing a cucumber([559.0, 626.0, 640.0, 835.0]), baguette([660.0, 704.0, 962.0, 944.0]), and sauce bottle sits to the right. Since the stapler is an office tool and the task requires placing such items into the steamer, and the object is already secured in the gripper over the correct container, the appropriate next action is to place the stapler into the steamer.",
  "tool_calls": [{
    "function": {
      "name": "robot_action",
      "arguments": {
        "action": "place",
        "label": "steamer",
        "point_2d": [516, 353]}}}]}
\end{lstlisting}

After each \texttt{robot\_action} call, a corresponding tool-result message is appended to the interaction history. The tool response provides only execution-level feedback, including whether the requested action is executable, whether the command has been dispatched, and the action arguments used for execution. It does not expose any task-specific checker information, such as task completion, progress, penalties, visibility scores, rewards, or evaluation outcomes. The policy must therefore determine subsequent actions from the updated visual observations and interaction history rather than from privileged task-state feedback. In the serialized history, each tool result is represented by a message with role \texttt{tool}, with the execution result stored in the \texttt{content} field:
\begin{lstlisting}
{"role": "tool",
  "content": {
    "command_dispatched": true,
    "action_executable": true,
    "action": "pick",
    "label": "stapler",
    "point_2d": [180, 457],
    "tool_response": "Found target from point and label. Pick command sent."}}

{"role": "tool",
  "content": {
    "command_dispatched": true,
    "action_executable": true,
    "action": "place",
    "label": "steamer",
    "point_2d": [516, 353],
    "tool_response": "Found target from point and label. Place command sent."}}
\end{lstlisting}
For multi-turn samples, the model receives the complete interaction history up to the current decision step, including previous visual observations, assistant responses, and tool results, together with the current observation. In contrast, a single-turn sample contains only the task instruction and the current observation, without any preceding interaction history. Checker outputs remain hidden from the policy in both settings and are used only for data validation, reward computation, and evaluation.

\subsection{Reward Instantiation}
\label{app:reward}
This section provides the detailed instantiation of the reward design. We specify the grounding, action, and outcome rewards, followed by their weighting and the semantic advantage shaping used during training.

\noindent\textbf{Grounding Reward.}
We parse each CoT grounding reference as a labeled 2D box of the form \texttt{label([}$x_1,y_1,x_2,y_2$\texttt{])}. For every predicted label, we resolve its corresponding scene actor by normalized actor-name and object-type matching. The ground-truth box for a resolved actor is the tight axis-aligned box of its visible segmentation mask in the current camera view, normalized to the $[0,1000]$ coordinate range. Predictions that cannot be resolved receive no credit.

Let $\mathcal{V}_t$ be the visible task-related objects, distractors, and containers at turn $t$, and let $\mathcal{U}_t\subseteq\mathcal{V}_t$ be the unique actors resolved from predicted labels. For each $i\in\mathcal{U}_t$, we retain only the prediction $\hat b_i$ with maximum IoU against the ground-truth box $b_i$ of actor $i$. This duplicate removal prevents repeated references to one object from receiving additional credit. We define grounding quality and coverage as
\begin{equation}
\operatorname{quality}_t=
\frac{1}{|\mathcal{U}_t|}\sum_{i\in\mathcal{U}_t}
\frac{|\hat b_i\cap b_i|}{|\hat b_i\cup b_i|},
\qquad
\operatorname{coverage}_t=
\min\!\left(1,\frac{|\mathcal{U}_t|}{|\mathcal{V}_t|}\right).
\end{equation}
The grounding score is $\operatorname{quality}_t \cdot \operatorname{coverage}_t$, and the reward is
\begin{equation}
r_t^{\mathrm{grd}}=
\operatorname{quality}_t \cdot \operatorname{coverage}_t.
\end{equation}
We set $r_t^{\mathrm{grd}}=0$ when $\mathcal{U}_t$ is empty. Quality measures localization accuracy over uniquely grounded actors, while coverage measures the fraction of visible candidate actors referenced by the rationale.

\noindent\textbf{Action Reward.}
The checker-based state score used by the action reward is
\begin{equation}
q(s)=\mathrm{progress}(s)-\mathrm{penalty}(s),
\end{equation}
where $\mathrm{progress}$ measures correct task advancement and $\mathrm{penalty}$ measures erroneous task changes. The complete action reward is
\begin{equation}
r_t^{\mathrm{act}}=
\begin{cases}
r_t^{\mathrm{exec}}+q(s_{t+1})-q(s_t), & a_t\in\{\texttt{pick},\texttt{place}\},\\
r_t^{\mathrm{exec}}+\mathrm{visibility}(s_{t+1})-\mathrm{visibility}(s_t), & a_t=\texttt{home},\\
r_t^{\mathrm{exec}}+\mathrm{success}(s_{t+1}), & a_t=\texttt{done}.
\end{cases}
\end{equation}
Here, $r_t^{\mathrm{exec}}$ is positive for a valid and executable tool call and negative for malformed calls, invalid points, or violated preconditions. The remaining terms use checker outputs to measure progress and error changes, visibility recovery, and successful termination.

\noindent\textbf{Outcome Reward.}
The outcome reward measures how much task progress a rollout achieves, how efficiently that progress is obtained, and whether the policy explicitly terminates a successfully completed task. For a rollout $\tau$, let $s_{\mathrm{init}}$ and $s_{\mathrm{final}}$ denote its initial and terminal states, and let $s^{\star}$ denote a checker-confirmed goal state. Using the checker state score $q(s)$ from Equation~\ref{eq:state-score}, we first normalize the progress achieved by the rollout relative to the total progress required from its initial state:
\begin{equation}
g(\tau)=
\operatorname{clip}\!\left(
\frac{q(s_{\mathrm{final}})-q(s_{\mathrm{init}})}
{q(s^{\star})-q(s_{\mathrm{init}})},0,1\right).
\end{equation}
Thus, $g(\tau)=0$ indicates no net progress beyond the initial state, $g(\tau)=1$ indicates that the goal-state score has been reached, and intermediate values give partial credit for incomplete but useful execution. The degenerate case $q(s_{\mathrm{init}})=q(s^{\star})$ means that the rollout already starts from a completed state and is handled separately. Because the initial state already attains the maximum state score, the rollout can only preserve this score or regress to a lower one. A lower terminal score, $q(s_{\mathrm{final}})<q(s_{\mathrm{init}})$, is treated as regression and receives $g(\tau)=0$. If the terminal score remains unchanged, $q(s_{\mathrm{final}})=q(s_{\mathrm{init}})$, the rollout receives $g(\tau)=1$ only if the policy outputs \texttt{done} to terminate the interaction, and $0$ otherwise.

We additionally reward interaction efficiency. Let $N_{\mathrm{env}}(\tau)$ denote the number of environment turns in the rollout and $N_{\max}$ the maximum allowed number of turns. We define
\begin{equation}
K=2\operatorname{round}\bigl(q(s^{\star})-q(s_{\mathrm{init}})\bigr)+1,
\end{equation}
which serves as a reference interaction length for completing the remaining task: each unit of task progress typically requires a \texttt{pick}--\texttt{place} pair, with one additional turn for the final \texttt{done} action. The efficiency factor is then
\begin{equation}
\mathrm{speed}(\tau)=\operatorname{clip}\!\left(
1-\frac{N_{\mathrm{env}}(\tau)-K}{\max(1,N_{\max}-K)},0,1\right).
\end{equation}
A rollout using at most $K$ turns receives the maximum efficiency factor, while additional turns progressively reduce this factor toward zero at the interaction limit. Importantly, efficiency is multiplied by task progress, so short rollouts that make little or no progress are not rewarded merely for terminating early.

\begin{equation}
\begin{aligned}
R^{\mathrm{success}}(\tau) &= 0.7g(\tau),\\
R^{\mathrm{step}}(\tau) &= 0.2g(\tau)\mathrm{speed}(\tau),\\
R^{\mathrm{done}}(\tau) &=
\begin{cases}
0.1, & \mathrm{traj\_success}=\texttt{True}\ \text{and}\ \mathrm{termination\_reason}=\texttt{done},\\
0, & \text{otherwise}.
\end{cases}
\end{aligned}
\end{equation}

We combine the three components to form the overall rollout-level outcome reward, which jointly captures task progress, execution efficiency, and correct termination. Our outcome reward is:
\begin{equation}
R^{\mathrm{out}}(\tau)
=
R^{\mathrm{success}}(\tau)
+R^{\mathrm{step}}(\tau)
+R^{\mathrm{done}}(\tau).
\end{equation}
Thus, partially successful rollouts retain their progress and efficiency components. We assign $r_t^{\mathrm{out}}=R^{\mathrm{out}}(\tau)$ to every turn in the rollout. The three terms respectively reward progress toward task completion, efficient execution with fewer environment interactions, and explicit termination after successful completion. The resulting rollout-level reward is assigned uniformly to every turn in the trajectory, i.e., each turn receives the same outcome reward computed from the final rollout result. This design preserves partial credit for trajectories that make meaningful progress, while assigning the maximum reward to trajectories that complete the task efficiently and terminate correctly.

\noindent\textbf{Reward Weights.}
We use $\lambda_{\mathrm{grd}}=0.25$, $\lambda_{\mathrm{act}}=0.50$, and $\lambda_{\mathrm{out}}=0.25$ in Equation~\ref{eq:turn-reward}.

\noindent\textbf{Semantic Advantage Shaping.}
For Equation~\ref{eq:semantic-advantage-shaping}, we use $\mu_{\mathrm{inv}}=-0.02$, $\mu_{\mathrm{deg}}=-0.01$, and $\mu_{\mathrm{imp}}=0.02$ for invalid, non-improving, and improving \texttt{pick}/\texttt{place} turns, respectively. Shaping is applied only to these two action types; for \texttt{home} and \texttt{done}, we retain $\widetilde{A}_t=A_t^{\mathrm{raw}}$.

\subsection{Training Configuration}
\label{app:training-configuration}
This section provides the training configurations for SFT and online RL. We describe the training data, optimization settings, rollout configuration, and computational resources used in each stage.

SFT uses 20,435 self-collected episodes to construct 65,000 training samples. Approximately one-third of the data consist of full multi-turn trajectories, while the remainder are individual turns sampled from nominal, intermediate, and recovery states. Each sample contains the task instruction, current observation, and, for multi-turn cases, the interaction history. The supervision target consists of a textual CoT rationale generated by Qwen3.7-Max and augmented with simulator-derived bounding-box annotations for grounding, followed by one valid tool call. For \texttt{pick} and \texttt{place}, the target additionally includes a 2D image-space point, which is converted into a 3D target through the environment interface. This SFT checkpoint serves as a cold start for online RL, providing the policy with basic instruction-following, grounded reasoning, and interaction capabilities.

The SFT training configuration is summarized in Table~\ref{tab:app-sft-training}. We train SFT for 8 epochs from Qwen3-VL-2B-Thinking with a learning rate $5\times10^{-6}$ and use the checkpoint at step 4,500. We use batch size 64, cosine scheduling with 3\% warmup, zero weight decay, maximum gradient norm 2, and bfloat16 precision. The vision encoder is frozen, while the multimodal projector and language model are trainable. We use gradient checkpointing, a maximum sequence length of 4096 tokens, image resolution bounds of 784--50,176 pixels, four data-loader workers, and DeepSpeed ZeRO-3.

\begin{table}[t]
\centering
\caption{SFT training configuration.}
\begin{tabular}{p{0.43\linewidth}p{0.43\linewidth}}
\toprule
Parameter & Setting \\
\midrule
Base model & Qwen3-VL-2B-Thinking \\
Epochs & 8 \\
Learning rate & $5\times10^{-6}$ \\
Batch size & 64 \\
Learning-rate schedule & Cosine \\
Warmup ratio & 3\% \\
Precision & bfloat16 \\
Trainable modules & Multimodal projector and language model \\
Vision encoder & Frozen \\
Maximum sequence length & 4096 tokens \\
Minimum image pixels & 784 \\
Maximum image pixels & 50,176 \\
Gradient checkpointing & Enabled \\
\bottomrule
\end{tabular}
\label{tab:app-sft-training}
\end{table}

SFT training used two NVIDIA A800 GPUs and took approximately 12 hours.

The online RL training configuration is summarized in Table~\ref{tab:app-rl-training}. Online RL starts from the SFT checkpoint and retains all preceding rendered observations and actions in the policy history throughout each multi-turn rollout. The checker supplies the rewards and semantic labels used by the RL objective. The maximum interaction budget is set to $2N_{\mathrm{target}}+4$, where $N_{\mathrm{target}}$ denotes the task-specific target count. This task-dependent budget is used identically for online RL rollouts and test episodes. The policy groups turns by discrete action and applies one turn-level GSPO objective across the task-length distribution. Environment transitions are not differentiated through, and no learned reward model or separate critic is used.

\begin{table}[t]
\centering
\caption{Online RL training configuration.}
\begin{tabular}{p{0.43\linewidth}p{0.43\linewidth}}
\toprule
Parameter & Setting \\
\midrule
Training batch size & 8 prompts \\
Rollouts per prompt & 6 \\
Trajectories per update & 48 \\
Loss aggregation & Sequence-mean token-mean \\
Learning rate & $2\times10^{-6}$ \\
Learning-rate schedule & Cosine decay \\
Warmup ratio & 2\% \\
Gradient clipping & 1.0 \\
PPO clip ratio & 0.2 \\
KL coefficient ($\beta$) & 0.02 \\
Maximum prompt length & 1,280 tokens \\
Maximum response length & 14,400 tokens \\
Maximum model length & 16,384 tokens \\
Rollout temperature & 0.9 \\
Maximum interaction budget & $2N_{\mathrm{target}}+4$ \\
Training updates & 200 \\
\bottomrule
\end{tabular}
\label{tab:app-rl-training}
\end{table}

For online RL, two NVIDIA A800 GPUs trained the policy while four NVIDIA RTX 3090 GPUs hosted the parallel interaction environments. The reported 200-update run took approximately 48 hours.

\subsection{Evaluation Metrics}
\label{app:metrics}
This section defines the evaluation metrics used throughout our experiments, including Success Rate (SR), Progress (PG), No-Penalty Rate (NPR), and Strict Success Rate (SSR).

For episode $i$, let $s_i\in\{0,1\}$ indicate completion of all required subtasks, $p_i\in\{0,1\}$ indicate that the action trace contains neither a grasp of a task-irrelevant object nor an object misplacement, and $d_i\in\{0,1\}$ indicate normal termination with \texttt{done}. Let $m_i$ and $M_i$ be the numbers of completed and total subtasks, respectively. For $N$ episodes, we compute the execution metrics as
\begin{equation}
\begin{aligned}
\mathrm{SR} &= \frac{1}{N}\sum_{i=1}^{N}s_i, &
\mathrm{PG} &= \frac{1}{N}\sum_{i=1}^{N}\frac{m_i}{M_i},\\
\mathrm{NPR} &= \frac{1}{N}\sum_{i=1}^{N}p_i, &
\mathrm{SSR} &= \frac{1}{N}\sum_{i=1}^{N}s_i p_i d_i.
\end{aligned}
\end{equation}
Thus, NPR measures measures the absence of self-induced execution errors during nominal and recovery rollouts, while SSR requires complete, error-free execution with normal termination.

For recovery evaluation, let $\mathcal D_{\mathrm{MR}}$,
$\mathcal D_{\mathrm{OR}}$, and $\mathcal D_{\mathrm{MTR}}$ denote the sets of
Manipulation Recovery, Occlusion Recovery, and Missing-Target Recovery episodes,
respectively. Each recovery episode is initialized from a controlled error state
containing the corresponding induced error. Recovery Rate measures whether the
model successfully corrects this induced error.

For MR and MTR, the initial state contains an erroneous object configuration
identified by the task-specific checker. We consider the induced error corrected
when, after the model's recovery execution, the checker no longer detects the
erroneous object relation introduced in the initial error state. For OR, each episode is initialized from a state in which the robot arm substantially occludes the workspace observation.
In ERR-Env, \texttt{home} is the designated recovery action for this condition:
executing it moves the robot away from the workspace and restores the occluded
observation. We therefore consider the induced occlusion corrected when the model
invokes \texttt{home} immediately after observing the occluded state.

Formally, for each recovery episode $i$, we define the recovery indicator
\begin{equation}
r_i^{\mathrm{rec}} =
\begin{cases}
1, & i\in\mathcal D_{\mathrm{MR}}\cup\mathcal D_{\mathrm{MTR}}
     \text{ and the induced object-state error is corrected},\\
1, & i\in\mathcal D_{\mathrm{OR}}
     \text{ and the action following the occluded observation is \texttt{home}},\\
0, & \text{otherwise}.
\end{cases}
\end{equation}
For each recovery type
$c\in\{\mathrm{MR},\mathrm{OR},\mathrm{MTR}\}$, the Recovery Rate is
\begin{equation}
\mathrm{RR}_{c}
=
\frac{1}{|\mathcal D_{c}|}
\sum_{i\in\mathcal D_{c}} r_i^{\mathrm{rec}}.
\end{equation}
Thus, Recovery Rate isolates the model's ability to resolve the specific error
present in the initial state, rather than requiring completion of the entire
remaining task. Recovery SSR complements this metric by applying the standard
Strict Success Rate criterion to recovery episodes, thereby measuring whether
the model can recover from the induced error and subsequently complete the
remaining task without additional execution errors. The reported Total Recovery
Rate and Total Recovery SSR are the arithmetic means of the corresponding MR,
OR, and MTR metrics.

\subsection{Additional Experimental Details}
\label{app:addition-exp}
This section provides additional details on the experimental setup and evaluation. We describe the baseline configurations and horizon-specific results, analyze tool-call format errors, and provide the zero-shot evaluation protocols and task configurations for the external benchmarks.

\label{app:baselines}
\noindent\textbf{Baselines.}
We consider three groups of baselines. The closed-source models include Qwen3.7-Plus~\citep{qwen2026qwen37plus}, GPT-5.6 Sol~\citep{openai2026gpt56sol}, and Qwen3.8-Max~\citep{qwen2026qwen38max}, which serve as strong general-purpose references for reasoning and tool use at substantially larger model scales. The general-purpose open-source VLMs include Qwen3-VL-235B-A22B~\citep{qwen3vl2025}, Qwen3-VL-2B-Thinking~\citep{qwen3vl2025} (the base model of ERR-VLM), and InternVL3.5-8B~\citep{internvl2025}, covering both large-capacity and compact pretrained VLMs. We further evaluate embodied VLMs, including RoboBrain2.5-8B~\citep{robobrain2026}, Hy-Embodied-VLM-1.0~\citep{hyembodied2026}, MiMo-Embodied-7B~\citep{hao2025mimoembodied}, and Embodied-R1.5~\citep{embodiedr15}, which are more directly targeted at interactive perception, grounding, planning, and action execution in physical environments. Notably, Embodied-R1.5 is additionally trained with error-related embodied data, making it a particularly relevant baseline for evaluating error recognition and recovery. For a controlled comparison, all prompted baselines use the same system prompt, rendered observations, complete interaction history, semantic tool definitions, OpenAI-compatible tool-call format, and 500-token thinking budget as \modelname{}. They are evaluated in a zero-shot setting without task-specific demonstrations, additional prompting heuristics, or parameter updates.

\subsubsection{Horizon-specific results}
The 1,503-task held-out \benchname{} test set is approximately balanced between short- and long-horizon episodes. Short-horizon episodes contain at most 10 ground-truth subtasks, whereas long-horizon episodes contain more than 10. Initial-state, intermediate-state, and error-state initializations are represented in balanced proportions; recovery episodes account for one-third of the test set and are evenly distributed across Manipulation Recovery, Occlusion Recovery, and Missing-Target Recovery. Table~\ref{tab:app-main-results-sl} reports the performance of the models in the main experiments separately on short- and long-horizon episodes, while Table~\ref{tab:app-ablations-sl} reports the corresponding short- and long-horizon results for the ablation experiments.

\begin{table*}[t]
\caption{Horizon-specific results on \benchname. Short- and long-horizon episodes are reported separately. SR: Success Rate; PG: Progress; NPR: No-Penalty Rate; SSR: Strict Success Rate. All entries are percentages.}
\centering
{\fontsize{9}{10.5}\selectfont
\setlength{\tabcolsep}{0pt}
\begin{tabular}{@{}>{\raggedright\arraybackslash}p{0.24\textwidth}@{\hspace{1pt}\vrule width 0.2pt\hspace{1pt}}*{4}{>{\centering\arraybackslash}p{0.0935\textwidth}}@{\hspace{1pt}\vrule width 0.2pt\hspace{1pt}}*{4}{>{\centering\arraybackslash}p{0.0935\textwidth}}@{}}
\toprule
\multicolumn{1}{l@{\hspace{1pt}\vrule width 0.2pt\hspace{1pt}}}{\multirow{2}{*}{Model}} & \multicolumn{4}{c@{\hspace{1pt}\vrule width 0.2pt\hspace{1pt}}}{Short-horizon} & \multicolumn{4}{c}{Long-horizon} \\
 & SR & PG & NPR & SSR & SR & PG & NPR & SSR \\
\midrule
\multicolumn{9}{l}{\textit{Closed-source models}} \\
GPT-5.6 Sol & \underline{40.35} & \underline{73.11} & 55.66 & 12.52 & \underline{41.62} & \underline{80.50} & 48.27 & 18.35 \\
Qwen3.8-Max & \textbf{65.38} & \textbf{83.25} & 59.25 & \textbf{30.23} & \textbf{44.95} & \textbf{82.06} & 47.47 & \textbf{24.20} \\
Qwen3.7-Plus & 21.30 & 52.57 & 57.26 & 3.33 & 15.03 & 52.58 & 52.53 & 7.45 \\
\midrule
\multicolumn{9}{l}{\textit{Open-source models}} \\
Qwen3-VL-235B-A22B & 36.35 & 71.95 & 39.68 & 12.92 & 16.76 & 65.85 & 42.02 & 7.58 \\
InternVL3.5-8B & 10.52 & 50.20 & 52.73 & 3.33 & 8.38 & 47.68 & 40.82 & 3.32 \\
Qwen3-VL-2B (Base) & 9.45 & 38.42 & \underline{69.37} & 3.33 & 6.65 & 42.10 & 63.30 & 0.80 \\
\midrule
\multicolumn{9}{l}{\textit{Embodied VLMs}} \\
RoboBrain2.5-8B & 15.58 & 50.70 & 44.47 & 5.59 & 9.97 & 47.72 & 40.82 & 2.53 \\
Hy-Embodied-VLM-1.0 & 6.26 & 36.80 & 68.58 & 3.46 & 7.58 & 43.96 & \textbf{67.02} & 4.65 \\
MiMo-Embodied-7B & 8.92 & 36.90 & \textbf{72.84} & 0.00 & 6.65 & 40.71 & \underline{65.82} & 1.73 \\
Embodied-R1.5 & 16.64 & 56.90 & 40.61 & 5.06 & 5.05 & 52.82 & 42.55 & 1.73 \\
\midrule
\multicolumn{9}{l}{\textit{Ours}} \\
\modelname{} (SFT) & 24.10 & 63.60 & 55.53 & 11.72 & 13.83 & 61.12 & 50.66 & 9.57 \\
\modelname{} (Full) & 38.88 & 71.45 & 63.65 & \underline{23.30} & 29.39 & 70.70 & 61.17 & \underline{18.48} \\
\bottomrule
\end{tabular}
}
\label{tab:app-main-results-sl}
\end{table*}

\begin{table*}[t]
\caption{Horizon-specific ablation results on \benchname. Short- and long-horizon episodes are reported separately. SR: Success Rate; PG: Progress; NPR: No-Penalty Rate; SSR: Strict Success Rate. All entries are percentages.}
\centering
{\fontsize{9}{10.5}\selectfont
\setlength{\tabcolsep}{0pt}
\begin{tabular}{@{}>{\raggedright\arraybackslash}p{0.24\textwidth}@{\hspace{1pt}\vrule width 0.2pt\hspace{1pt}}*{4}{>{\centering\arraybackslash}p{0.0935\textwidth}}@{\hspace{1pt}\vrule width 0.2pt\hspace{1pt}}*{4}{>{\centering\arraybackslash}p{0.0935\textwidth}}@{}}
\toprule
\multirow{2}{*}{Method} & \multicolumn{4}{c@{\hspace{1pt}\vrule width 0.2pt\hspace{1pt}}}{Short-horizon} & \multicolumn{4}{c}{Long-horizon} \\
 & SR & PG & NPR & SSR & SR & PG & NPR & SSR \\
\midrule
\multicolumn{9}{l}{\textit{SFT and full RL}} \\
\modelname{} (SFT) & 24.10 & 63.60 & 55.53 & 11.72 & 13.83 & 61.12 & 50.66 & 9.57 \\
\modelname{} (Full) & 38.88 & 71.45 & \underline{63.65} & \textbf{23.30} & 29.39 & 70.70 & \textbf{61.17} & \underline{18.48} \\
\midrule
\multicolumn{9}{l}{\textit{Recovery-data Ablations}} \\
w/o Recovery Data in RL & 35.02 & 69.20 & 55.53 & 16.64 & 15.03 & 69.07 & 50.00 & 11.70 \\
\midrule
\multicolumn{9}{l}{\textit{Reward Ablations}} \\
w/o Grounding Reward & 35.02 & 71.06 & 60.59 & 18.38 & \underline{30.05} & 73.64 & 51.73 & 14.23 \\
w/o Action Reward & 38.88 & 74.71 & 48.34 & 10.12 & \textbf{34.18} & 73.71 & 38.30 & 9.97 \\
w/o Penalty Feedback & 37.28 & 72.95 & 47.27 & 3.86 & 26.73 & 71.94 & 39.23 & 1.73 \\
w/o Outcome Reward & 39.41 & 70.82 & 54.99 & 12.78 & 21.68 & 71.84 & 44.15 & 10.77 \\
\midrule
\multicolumn{9}{l}{\textit{Group Ablations}} \\
Batch (All) & \textbf{42.74} & \textbf{78.44} & 61.12 & \underline{20.51} & 29.12 & \underline{76.12} & 56.65 & \textbf{19.15} \\
Sample (All) & \underline{41.15} & 75.50 & 53.93 & 7.72 & 22.47 & 75.82 & 47.47 & 10.77 \\
Batch (Action) & 39.95 & \underline{78.38} & 51.66 & 13.85 & 23.27 & \textbf{77.06} & 44.15 & 8.38 \\
\midrule
\multicolumn{9}{l}{\textit{RL Component Ablations}} \\
w/o Advantage Shaping & 24.50 & 64.46 & 51.13 & 10.52 & 29.92 & 71.82 & 51.73 & 10.77 \\
\midrule
\multicolumn{9}{l}{\textit{Algorithm Ablations}} \\
Turn-GRPO & \underline{41.15} & 76.45 & 54.46 & 7.72 & 22.47 & 73.33 & 53.32 & 9.97 \\
Trajectory-GSPO & 14.38 & 49.82 & \textbf{70.04} & 11.05 & 7.45 & 52.37 & \underline{60.77} & 5.05 \\
Trajectory-GRPO & 17.84 & 57.25 & 63.38 & 12.25 & 9.97 & 58.95 & 56.65 & 6.65 \\
\bottomrule
\end{tabular}
}
\label{tab:app-ablations-sl}
\end{table*}

\noindent\textbf{Horizon-Specific Results on ERR-Bench.}
Table~\ref{tab:app-main-results-sl} shows that long-horizon episodes are consistently more challenging for most existing models. In particular, success-oriented metrics such as SR and SSR generally decrease as the horizon grows, reflecting the increased difficulty of maintaining correct execution over longer interaction sequences. At the same time, several models retain relatively high PG on long-horizon episodes despite much lower SR and SSR. For example, Qwen3.8-Max achieves 82.06\% PG on long-horizon tasks, while its SR and SSR are only 44.95\% and 24.20\%, respectively. A similar gap appears for other baselines. This discrepancy indicates that making substantial task progress does not necessarily translate into reliable execution: models may complete many intermediate subtasks but still fail due to accumulated grounding, planning, or action errors before reaching the final goal.

Our training improves performance consistently across both horizon regimes. Compared with the SFT policy, the full \modelname{} increases SR from 24.10\% to 38.88\% on short-horizon episodes and from 13.83\% to 29.39\% on long-horizon episodes. The gains are also consistent in PG, NPR, and SSR, with SSR improving from 11.72\% to 23.30\% on short-horizon tasks and from 9.57\% to 18.48\% on long-horizon tasks. These results show that the benefits of our training are not restricted to short interactions, but extend to longer-horizon settings where successful execution requires maintaining reliable decisions over a larger number of steps.

\noindent\textbf{Horizon-Specific Ablation Results on ERR-Bench.}
The group ablations show that the grouping strategy substantially affects the trade-off between task progress and reliable execution. Batch (All) achieves the strongest short-horizon SR and PG, reaching 42.74\% and 78.44\%, respectively, and also obtains the highest long-horizon SSR of 19.15\% among the group variants. However, these gains do not consistently translate into more reliable behavior, as its NPR remains below that of the full model on both horizons. This discrepancy is more pronounced for Sample (All) and Batch (Action). Both variants achieve relatively high PG, with Batch (Action) reaching 78.38\% on short-horizon episodes and 77.06\% on long-horizon episodes, while exhibiting substantially lower NPR and SSR. In particular, Batch (Action) achieves only 13.85\% SSR on short-horizon episodes and 8.38\% on long-horizon episodes despite its high PG. These results suggest that broader or action-focused grouping can effectively encourage intermediate task progress, but does not necessarily provide sufficiently discriminative learning signals for avoiding execution errors and reliably completing the full task. In comparison, the full grouping design yields a more balanced performance across SR, NPR, and SSR, particularly on long-horizon episodes where errors accumulated over multiple interaction steps become more consequential.

The algorithm ablations reveal two complementary effects: the granularity of policy updates within each generated response and the granularity of credit assignment across interaction turns. Comparing AGT-GSPO with Turn-GRPO, both methods retain turn-level rewards and therefore preserve local credit assignment, but they differ in how policy updates are weighted. GRPO applies token-level importance ratios and clipping, allowing different tokens within the same response to receive different update weights. In contrast, GSPO computes a length-normalized sequence-level importance ratio and applies a shared clipping decision to the response as a whole. This distinction is particularly relevant in our setting, where each turn receives a unified reward reflecting the quality of the resulting grounding and action. Sequence-level weighting better aligns the policy update with this turn-level reward, whereas token-wise weighting can introduce inconsistent updates among tokens that jointly determine the same structured action. This helps explain why Turn-GRPO can still achieve strong task advancement, with 41.15\% SR and 76.45\% PG on short-horizon episodes, yet obtains only 7.72\% SSR, substantially below the 23.30\% of the full ERR-VLM model. The same pattern persists on long-horizon episodes: Turn-GRPO maintains 73.33\% PG but reaches only 22.47\% SR and 9.97\% SSR. These results suggest that token-level updates can learn behaviors that frequently advance the task, while the more coherent response-level updates of GSPO are better suited to translating such progress into reliable execution.

The gap between turn-level and trajectory-level optimization is substantially larger, reflecting a different limitation. Under turn-level optimization, each interaction turn is evaluated separately, so the policy can directly associate the quality of a grounding or action decision with the tokens that produced that decision. Trajectory-level optimization instead assigns a single reward and advantage to the entire multi-turn interaction. Consequently, successful intermediate decisions and erroneous turns receive the same global credit signal, making it difficult to identify which decisions should be reinforced or suppressed. This credit-assignment ambiguity becomes increasingly severe as the horizon grows because more heterogeneous decisions are compressed into the same trajectory-level signal. Consistent with this interpretation, Trajectory-GSPO reaches only 14.38\% SR on short-horizon episodes and further decreases to 7.45\% on long-horizon episodes, while Trajectory-GRPO obtains 17.84\% and 9.97\% SR, respectively. Their PG values are likewise substantially below those of the turn-level variants. Notably, Trajectory-GSPO still achieves 70.04\% NPR on short-horizon episodes and 60.77\% on long-horizon episodes despite its low SR and PG. One plausible explanation is that coarse trajectory-level supervision can favor conservative behaviors that avoid penalized actions without providing sufficiently localized signals for making the sequence of decisions required for task completion. Taken together, these results indicate that the main benefit comes from preserving turn-level credit assignment, while GSPO further improves the consistency of policy updates by aligning its sequence-level optimization unit with the reward assigned to each turn.

\subsubsection{Tool-call format analysis}
We additionally analyze tool-call formatting behavior to verify that the main evaluation results are not primarily explained by compatibility with the tool-call interface. We distinguish three types of format errors: a missing tool call, where the required action call is omitted entirely; a parse failure, where a generated call cannot be parsed; and an invalid tool/arguments error, where the call is syntactically parseable but violates the tool schema or argument constraints. Table~\ref{tab:tool-call-errors} reports these errors at the turn level. This view captures how often formatting failures occur during interaction, but does not by itself indicate whether a trajectory becomes non-executable, since a model may return to a valid tool-call format in a subsequent turn.

\begin{table*}[t]
\centering
{\fontsize{8}{9.5}\selectfont
\caption{Tool-call format errors observed during the main-result evaluation. Counts are error turns rather than unique trajectories.}
\begin{tabular}{lrrrr}
\toprule
Model & Missing Tool Call & Parse Failure & Invalid Tool/Arguments & Total Error Turns \\
\midrule
GPT-5.6 Sol & 5 & 0 & 0 & 5 \\
Qwen3.8-Max & 744 & 51 & 29 & 824 \\
Qwen3.7-Plus & 0 & 0 & 0 & 0 \\
Qwen3-VL-235B-A22B & 0 & 0 & 3 & 3 \\
InternVL3.5-8B & 2434 & 1754 & 271 & 4459 \\
Qwen3-VL-2B (Base) & 3907 & 314 & 36 & 4257 \\
RoboBrain2.5-8B & 0 & 105 & 503 & 608 \\
Hy-Embodied-VLM-1.0 & 0 & 0 & 26 & 26 \\
MiMo-Embodied-7B & 202 & 699 & 97 & 998 \\
Embodied-R1.5 & 85 & 0 & 15 & 100 \\
\modelname{} (SFT) & 55 & 0 & 0 & 55 \\
\modelname{} (Full) & 19 & 0 & 0 & 19 \\
\bottomrule
\end{tabular}}
\label{tab:tool-call-errors}
\end{table*}

To distinguish transient format errors from failures that prevent continued execution, Table~\ref{tab:tool-call-trajectories} summarizes their trajectory-level consequences. An Error Trajectory contains at least one format-error turn. State-Changing Executed Trajectories are error trajectories in which the model subsequently produces at least one executable action that changes the environment state. Later Clean-Format Turn counts error trajectories in which the model subsequently returns to a valid and executable tool call. In contrast, All Remaining Turns Malformed identifies trajectories for which every subsequent response remains malformed after a format error, preventing further execution. We report the corresponding Execution-Affected Rate over all evaluation episodes. 

\begin{table*}[t]
\centering
{\fontsize{7.5}{8.5}\selectfont
\setlength{\tabcolsep}{1pt}
\caption{Trajectory-level consequences of tool-call format errors in the main-result evaluation. Execution-Affected Rate is All Remaining Turns Malformed divided by 1,503 episodes.}
\begin{tabular}{@{}>{\raggedright\arraybackslash}p{0.19\textwidth}>{\centering\arraybackslash}p{0.10\textwidth}>{\centering\arraybackslash}p{0.20\textwidth}>{\centering\arraybackslash}p{0.16\textwidth}>{\centering\arraybackslash}p{0.20\textwidth}>{\centering\arraybackslash}p{0.12\textwidth}@{}}
\toprule
Model & \shortstack{Error\\trajectories} & \shortstack{State-changing\\executed trajectories} & \shortstack{Later clean-\\format turn} & \shortstack{All remaining\\turns malformed} & \shortstack{Execution-\\affected rate} \\
\midrule
GPT-5.6 Sol & 5 & 0 & 0 & 5 & 0.3\% \\
Qwen3.8-Max & 499 & 477 & 489 & 10 & 0.7\% \\
Qwen3.7-Plus & 0 & 0 & 0 & 0 & 0.0\% \\
Qwen3-VL-235B-A22B & 2 & 2 & 2 & 0 & 0.0\% \\
InternVL3.5-8B & 1304 & 1127 & 1244 & 60 & 4.0\% \\
Qwen3-VL-2B (Base) & 1351 & 404 & 1351 & 0 & 0.0\% \\
RoboBrain2.5-8B & 77 & 35 & 39 & 38 & 2.5\% \\
Hy-Embodied-VLM-1.0 & 24 & 12 & 24 & 0 & 0.0\% \\
MiMo-Embodied-7B & 368 & 21 & 309 & 59 & 3.9\% \\
Embodied-R1.5 & 19 & 12 & 14 & 5 & 0.3\% \\
\modelname{} (SFT) & 54 & 47 & 49 & 5 & 0.3\% \\
\modelname{} (Full) & 18 & 16 & 18 & 0 & 0.0\% \\
\bottomrule
\end{tabular}}
\label{tab:tool-call-trajectories}
\end{table*}

The two tables reveal a clear distinction between the frequency of format errors and their trajectory-level impact. Although some models, particularly InternVL3.5-8B and Qwen3-VL-2B, produce format errors in many turns and trajectories, these errors are predominantly transient. A trajectory containing a malformed turn can often return to a valid executable tool call and continue interacting with the environment. For example, all 1,351 Qwen3-VL-2B trajectories containing format errors subsequently recover a valid tool-call format. In contrast, the more consequential failure mode in which all remaining turns stay malformed and no further action can be executed is rare, affecting at most 60 of the 1,503 evaluation episodes (4.0\%) for any evaluated model. Our full model exhibits 18 error trajectories, all of which subsequently return to a valid executable format. These results show that a local tool-call formatting error does not generally render an entire trajectory non-executable. Although malformed turns may consume part of the interaction budget, persistent format breakdowns affect only a small fraction of the evaluation set, suggesting that differences in tool-call compatibility do not substantially confound the overall task-execution comparison.

\subsubsection{Details on Ablation}

This section provides additional details on the ablation experiments, including the ablation configurations and detailed analysis of the grouping strategy and algorithm variants.

\paragraph{Ablation Configurations.}

We describe the configurations of the ablation studies. We first specify the recovery-data, reward, penalty-feedback, and semantic-advantage ablations, and then describe the Turn-GRPO and trajectory-level variants used to compare different optimization and credit-assignment strategies.

All ablations start from the same SFT checkpoint and use the same rollout budget, optimizer, and training schedule as the full model unless stated otherwise.

\noindent\textbf{$\bullet$~Recovery-Data and Reward Ablations.}
For \emph{w/o recovery data}, online RL is initialized only from nominal and intermediate snapshots; recovery snapshots are removed from the RL sampling pool. The three reward ablations remove one term from Equation~\ref{eq:turn-reward} while leaving the remaining terms and their weights unchanged:
\begin{equation}
\begin{aligned}
r_t^{\mathrm{w/o\ grd}}&=\lambda_{\mathrm{act}}r_t^{\mathrm{act}}+\lambda_{\mathrm{out}}r_t^{\mathrm{out}},\\
r_t^{\mathrm{w/o\ act}}&=\lambda_{\mathrm{grd}}r_t^{\mathrm{grd}}+\lambda_{\mathrm{out}}r_t^{\mathrm{out}},\\
r_t^{\mathrm{w/o\ out}}&=\lambda_{\mathrm{grd}}r_t^{\mathrm{grd}}+\lambda_{\mathrm{act}}r_t^{\mathrm{act}}.
\end{aligned}
\label{eq:reward-ablations}
\end{equation}
Thus, \emph{w/o grounding reward} retains executable transition and rollout-outcome feedback, \emph{w/o action reward} retains grounding and rollout-outcome feedback, and \emph{w/o outcome reward} retains grounding and checker-verified turn feedback.

\noindent\textbf{$\bullet$~Penalty and Advantage Ablations.}
For \emph{w/o penalty feedback}, we replace the checker state score in Equation~\ref{eq:state-score} with $q^{-}(s)=\mathrm{progress}(s)$. Consequently, the \texttt{pick}/\texttt{place} reward becomes $r_t^{\mathrm{exec}}+\mathrm{progress}(s_{t+1})-\mathrm{progress}(s_t)$; \texttt{home} and \texttt{done} rewards are unchanged. For \emph{w/o semantic advantage shaping}, we directly optimize with $A_t^{\mathrm{raw}}$ from Equation~\ref{eq:raw-adv} in place of $\widetilde{A}_t$.

\noindent\textbf{$\bullet$~Group Ablations.}
These variants change only the comparison set used to compute $A_t^{\mathrm{raw}}$; they retain the full turn-level reward, semantic advantage shaping, and full turn-level objective. For sample $i$, let $\mathcal{O}_i=\bigcup_{k=1}^{K}\tau_i^{(k)}$ denote the set of all turns in its $K$ rollout trajectories, and let $\mathcal{B}$ denote the current update batch. Each variant uses
\begin{equation}
A_t^{\mathrm{raw}}=
\frac{r_t-\mu_{\mathcal{G}_t}}
{\sigma_{\mathcal{G}_t}+\delta},
\end{equation}
with the following comparison sets. For the full method,
\begin{equation}
\mathcal{G}_t^{\mathrm{full}}=
\begin{cases}
\{u\in\mathcal{O}_i\mid a_u=a_t\},
& a_t\in\{\texttt{pick},\texttt{place}\},\\
\{u\in\mathcal{B}\mid a_u=a_t\},
& a_t\in\{\texttt{home},\texttt{done}\}.
\end{cases}
\end{equation}
Thus, \texttt{pick}/\texttt{place} use same-type turns from the multiple rollouts of one sample, whereas the sparse \texttt{home}/\texttt{done} actions use same-type turns from the update batch. The three group ablations instead use
\begin{equation}
\mathcal{G}_t^{\mathrm{Sample\ (All)}}=\mathcal{O}_i,
\end{equation}
\begin{equation}
\mathcal{G}_t^{\mathrm{Batch\ (All)}}=\mathcal{B},
\end{equation}
and
\begin{equation}
\mathcal{G}_t^{\mathrm{Batch\ (Action)}}=
\{u\in\mathcal{B}\mid a_u=a_t\}.
\end{equation}
Here, \emph{All} means that every turn in the corresponding sample-level or batch-level set belongs to one group, without separating turns by semantic action type. \emph{Sample} therefore compares all turns from the multiple rollout trajectories of one sample, while \emph{Batch} compares turns pooled across the current update batch.

\noindent\textbf{$\bullet$~Turn-GRPO.}
Turn-GRPO retains the full turn-level reward, action-conditioned grouping, and semantic advantage shaping of AGT-GSPO. It replaces GSPO's response-level ratio with token-level ratios. For token $j$ in response $y_t$, let
\begin{equation}
\rho_{t,j}(\theta)=
\frac{\pi_\theta(y_{t,j}\mid y_{t,<j},I_t,L,H_t)}
{\pi_{\mathrm{old}}(y_{t,j}\mid y_{t,<j},I_t,L,H_t)}.
\end{equation}
Its clipped objective is
\begin{equation}
\mathcal{L}_{\mathrm{Turn\text{-}GRPO}}(\theta)=
\mathbb{E}_{\tau,t}\!\left[
\frac{1}{|y_t|}\sum_{j=1}^{|y_t|}
\min\!\left(\rho_{t,j}\widetilde{A}_t,
\operatorname{clip}(\rho_{t,j},1-\epsilon,1+\epsilon)\widetilde{A}_t\right)
\right]-\beta\mathcal{L}_{\mathrm{KL}}.
\label{eq:turn-grpo}
\end{equation}

\noindent\textbf{$\bullet$~Trajectory-Level Variants.}
Trajectory-GSPO and Trajectory-GRPO use only the outcome reward $R^{\mathrm{out}}(\tau)$ defined in Equation~\ref{eq:outcome-reward}; grounding and action rewards are removed. For rollout group $\mathcal{O}$ generated from the same initial context, they compare outcome rewards across trajectories and assign one shared advantage to every turn in a trajectory:
\begin{equation}
A_\tau=\frac{R^{\mathrm{out}}(\tau)-\mu_{\mathcal{O}}^{\mathrm{out}}}
{\sigma_{\mathcal{O}}^{\mathrm{out}}+\delta},
\qquad A_t=A_\tau\quad \forall t\in\tau,
\label{eq:trajectory-advantage}
\end{equation}
where $\mu_{\mathcal{O}}^{\mathrm{out}}$ and $\sigma_{\mathcal{O}}^{\mathrm{out}}$ are computed over trajectory outcomes in $\mathcal{O}$. Let $Y_\tau$ be the concatenation of all response tokens in trajectory $\tau$. Trajectory-GSPO optimizes the whole trajectory with
\begin{equation}
\rho_\tau(\theta)=
\left(
\frac{\pi_\theta(Y_\tau\mid\tau)}
{\pi_{\mathrm{old}}(Y_\tau\mid\tau)}
\right)^{1/|Y_\tau|},
\label{eq:trajectory-gspo-ratio}
\end{equation}
with objective
\begin{equation}
\mathcal{L}_{\mathrm{Traj\text{-}GSPO}}(\theta)=
\mathbb{E}_{\tau}\!\left[
\min\!\left(\rho_\tau A_\tau,
\operatorname{clip}(\rho_\tau,1-\epsilon,1+\epsilon)A_\tau\right)
\right]-\beta\mathcal{L}_{\mathrm{KL}}.
\label{eq:trajectory-gspo}
\end{equation}
Trajectory-GRPO instead applies a token-level ratio to the same shared trajectory advantage:
\begin{equation}
\mathcal{L}_{\mathrm{Traj\text{-}GRPO}}(\theta)=
\mathbb{E}_{\tau}\!\left[
\frac{1}{|Y_\tau|}\sum_{j=1}^{|Y_\tau|}
\min\!\left(\rho_{\tau,j}A_\tau,
\operatorname{clip}(\rho_{\tau,j},1-\epsilon,1+\epsilon)A_\tau\right)
\right]-\beta\mathcal{L}_{\mathrm{KL}},
\label{eq:trajectory-grpo}
\end{equation}
where $\rho_{\tau,j}$ is the token-level importance ratio for token $j$ in $Y_\tau$. These variants isolate the effect of trajectory-level outcome comparison and shared credit assignment from action-grouped turn-level learning.

\paragraph{Analysis of the Grouping Strategy.}
We further analyze how different grouping strategies affect advantage estimation and execution performance, focusing on the trade-off between comparison stability and the semantic comparability of grouped turns.

The group ablations show that effective advantage estimation depends on both statistical stability and the comparability of grouped turns. Sample (All) keeps turns within the same sample but mixes different action types. Since different actions have distinct reward semantics and distributions, their relative advantages can reflect action-type differences rather than execution quality. Batch (Action) avoids this cross-action mismatch by grouping the same action type, but compares turns across different samples, where task stage, state difficulty, and remaining progress can differ substantially. Notably, both variants achieve higher PG than the full method while showing much lower All-Task SSR and Recovery SSR. This pattern suggests that these coarser groups can still favor actions that improve the current task state, but provide less precise relative supervision for selecting actions that consistently lead to complete execution.

Batch (All) relaxes both grouping constraints but still performs comparably on PG and better on aggregate recovery metrics. One possible explanation is that its larger comparison pool yields more stable normalization statistics, preserving a strong coarse optimization signal despite the heterogeneous samples. However, this stability comes at the cost of semantic precision: mixing different action types and task contexts makes the resulting advantages reflect broad outcome differences across heterogeneous turns, rather than fine-grained distinctions among decisions made under comparable conditions. This coarse signal remains effective for promoting progress and recovery, but is less informative for optimizing the full interaction sequence toward strict task completion. Consistent with this interpretation, Batch (All) achieves a Recovery Rate of 63.07, compared with 65.47 for the full method, while its Recovery SSR drops more noticeably from 22.16 to 18.16.

In contrast, our within-sample, same-action grouping for pick and place controls both action semantics and task context, making reward differences more directly attributable to decision quality. This finer comparison produces a more informative relative signal for distinguishing actions that merely make local progress from those that better support successful completion. Although the full method does not maximize PG, it achieves the strongest All-Task SSR and Recovery SSR, indicating that more comparable grouping better converts verified local outcomes into reliable task-level execution and recovery.

\paragraph{Analysis of the Algorithms.}

\begin{table*}[t]
\captionsetup{
    justification=raggedright,
    singlelinecheck=false
}
\caption{More ablation results on \benchname. Metrics follow Table~\ref{tab:main-results}. All entries are percentages.}
\centering
{\fontsize{8}{9.5}\selectfont
\setlength{\tabcolsep}{0pt}
\begin{tabular}{@{}>{\raggedright\arraybackslash}p{0.22\textwidth}@{\hspace{1pt}\vrule width 0.2pt\hspace{1pt}}*{4}{>{\centering\arraybackslash}p{0.0636\textwidth}}@{\hspace{1pt}\vrule width 0.2pt\hspace{1pt}}*{4}{>{\centering\arraybackslash}p{0.0636\textwidth}}@{\hspace{1pt}\vrule width 0.2pt\hspace{1pt}}*{4}{>{\centering\arraybackslash}p{0.0636\textwidth}}@{}}
\toprule
\multirow{2}{*}{Method} & \multicolumn{4}{c@{\hspace{1pt}\vrule width 0.2pt\hspace{1pt}}}{All Tasks} & \multicolumn{4}{c@{\hspace{1pt}\vrule width 0.2pt\hspace{1pt}}}{Recovery Rate} & \multicolumn{4}{c}{Recovery SSR} \\
 & SR & PG & NPR & SSR & MR & OR & MTR & Total & MR & OR & MTR & Total \\
\midrule
\multicolumn{13}{l}{\textit{SFT and full RL}} \\
\modelname{} (SFT) & 18.96 & 62.36 & 53.09 & 10.64 & 26.95 & 40.72 & 59.88 & 42.52 & 13.17 & 4.19 & 8.38 & 8.58 \\
\modelname{} (Full) & 34.14 & 71.08 & 62.41 & 20.89 & 40.12 & 89.22 & 67.07 & 65.47 & 26.35 & 24.55 & 15.57 & 22.16 \\
\midrule
\multicolumn{13}{l}{\textit{Algorithm Ablations}} \\
Turn-GRPO & 31.81 & 74.89 & 53.89 & 8.85 & 45.51 & 83.83 & 48.50 & 59.28 & 13.77 & 10.18 & 9.58 & 11.18 \\
Trajectory-GSPO (Outcome) & 10.92 & 51.10 & 65.41 & 8.05 & 17.37 & 56.29 & 67.66 & 47.11 & 7.78 & 1.20 & 7.78 & 5.59 \\
Trajectory-GRPO (Outcome) & 13.91 & 58.10 & 60.02 & 9.45 & 23.95 & 59.88 & 65.27 & 49.70 & 6.59 & 1.20 & 7.19 & 4.99 \\
Trajectory-GSPO (Sum-of-Turn) & 8.91 & 45.35 & 55.95 & 7.05 & 16.17 & 52.10 & 58.08 & 42.12 & 6.59 & 0.60 & 7.78 & 4.99 \\
Trajectory-GSPO (Avg-of-Turn) & 15.90 & 57.10 & 64.41 & 10.11 & 25.75 & 60.48 & 61.08 & 49.10 & 10.18 & 2.40 & 9.58 & 7.39 \\
\bottomrule
\end{tabular}
}
\label{tab:ablations-more}
\vspace{-12pt}
\end{table*}

We provide a more detailed analysis of the algorithm ablations in Table~\ref{tab:ablations-more}. Trajectory-GSPO (Outcome) and Trajectory-GRPO (Outcome) correspond to Trajectory-GSPO and Trajectory-GRPO in Table~\ref{tab:ablations}, respectively. We further include trajectory-level variants that aggregate dense turn rewards by summation or averaging, providing additional comparisons between sparse outcome supervision and dense reward aggregation at the trajectory level. Due to space constraints, and because dense turn-level feedback is not naturally aligned with trajectory-level credit assignment, these additional variants are reported and analyzed in the appendix.

For a rollout $\tau$ with $T_\tau$ interaction turns, let $r_t$ denote the reward at turn $t$. We construct two trajectory-level dense reward variants by aggregating the turn-level rewards into a single scalar. Trajectory-GSPO (Sum-of-Turn) uses the sum of all turn-level rewards:
\begin{equation}
R_{\mathrm{sum}}(\tau)=\sum_{t=1}^{T_\tau} r_t.
\label{eq:traj-sum-turn}
\end{equation}
Trajectory-GSPO (Avg-of-Turn) instead uses their average:
\begin{equation}
R_{\mathrm{avg}}(\tau)=\frac{1}{T_\tau}\sum_{t=1}^{T_\tau} r_t.
\label{eq:traj-avg-turn}
\end{equation}
In both variants, the aggregated reward is treated as a single trajectory-level signal and used to compute one advantage for the entire rollout under the same Trajectory-GSPO optimization framework. The sum variant accumulates rewards over all interaction turns and is therefore sensitive to trajectory length, whereas the average variant normalizes by the number of turns to reduce this dependence.

We first compare AGT-GSPO with Turn-GRPO to isolate the effect of the policy optimization objective under the same turn-level formulation. Turn-GRPO attains slightly higher PG and MR Recovery Rate, but its strict completion metrics are substantially lower: All-Task SSR drops from 20.89 to 8.85, and Recovery SSR from 22.16 to 11.18. This indicates that Turn-GRPO can still learn behaviors that make local progress or recover individual states, but does not translate these improvements into equally reliable multi-turn completion. In contrast, GSPO applies the policy update at the sequence level, keeping each reasoning-and-action output as a coherent optimization unit. This is better aligned with multi-turn interaction, where the quality of a turn depends on the consistency of the entire generated sequence rather than on independently weighted tokens. The stronger SSR and Recovery SSR of the full ERR-VLM therefore suggest that sequence-level optimization provides a more suitable and stable update for learning reliable multi-turn execution.

We next compare turn-level and trajectory-level learning. Both Trajectory-GSPO (Outcome) and Trajectory-GRPO (Outcome), which assign a single outcome reward to the entire rollout, perform substantially worse than the full turn-level method, indicating that trajectory-level supervision is too coarse for effective credit assignment over multi-turn decisions. Trajectory-GRPO (Outcome) performs slightly better than Trajectory-GSPO (Outcome) on several aggregate metrics, but the gap is small and both remain far behind the full method, especially on strict recovery success. This suggests that changing the trajectory-level policy objective alone does not resolve the main limitation of assigning the same sequence-level supervision to all decisions in the rollout.

We further evaluate Trajectory-GSPO (Sum-of-Turn) and Trajectory-GSPO (Avg-of-Turn), which aggregate dense turn-level rewards into a single trajectory reward by summation and averaging, respectively. Trajectory-GSPO (Avg-of-Turn) performs slightly better than Trajectory-GSPO (Outcome) on several metrics, including All-Task SR and SSR and total Recovery SSR, while Trajectory-GSPO (Sum-of-Turn) is generally worse. However, both variants remain substantially below the full method. Even the stronger Avg-of-Turn variant reaches only 10.11 All-Task SSR and 7.39 Recovery SSR, compared with 20.89 and 22.16 for the full method. Aggregating turn-level feedback into a single trajectory-level scalar removes its temporal specificity, making it harder to identify which individual decisions contributed to progress, recovery, or failure. In contrast, the full method retains feedback at the interaction turn where it is observed, providing more direct credit assignment for multi-turn execution and recovery.

\subsubsection{Zero-shot Protocol}
For transfer evaluation, RoboBrain2.5-8B, the SFT policy, and the full \modelname{} policy are evaluated directly in RMBench, RoboVoLo, and real-world robotic settings without any environment-specific adaptation. All model parameters are kept fixed throughout evaluation, and no additional demonstrations, gradient updates, or reward tuning are introduced for the target environments. To support a consistent action space across different platforms, we implement deterministic environment-specific adapters that translate the unified model actions into the native control interfaces of each benchmark or robot. These adapters contain no learned parameters and do not alter the model outputs or provide additional task information. Each task is evaluated over 10 independent trials, allowing us to assess whether the planning and recovery behaviors learned in the training environment transfer to unseen simulators and real-world scenarios.

\noindent\textbf{RMBench.}
RMBench (Memory-Dependent Robotic Manipulation Benchmark) is a robotic manipulation benchmark for evaluating long-term memory. It systematically studies how robot policies store, retrieve, and use historical information in partially observable, long-horizon tasks. Unlike conventional manipulation benchmarks, which primarily emphasize short-term action execution or completion of long sequences, RMBench focuses on tasks where the current observation alone is insufficient for correct decision-making and successful completion requires information from earlier observations. We report the benchmark's native progress metric for each task.

The three evaluated tasks are:
\begin{itemize}\setlength{\itemsep}{2pt}\setlength{\parskip}{0pt}\setlength{\parsep}{0pt}
    \item \textbf{Observe and Pick Up:} Tests visual observation and short-term memory. Five categories of candidate objects are arranged on the tabletop. The robot first observes a target example on a shelf; after the example is occluded, the policy must use the observation to locate and grasp an object of the same category among the candidates. Success requires completing the observation stage and grasping a candidate that matches the example category.
    \item \textbf{Cover Blocks:} Tests multi-stage state understanding and order constraints. The task contains three colored blocks and covers. The policy must first move the three covers over their blocks in order, then remove the covers to reveal the blocks in a per-episode randomized color order. The environment maintains a stage-state machine; selecting an incorrect cover at any stage sets a failure flag, so subsequent actions cannot recover success.
    \item \textbf{Place Block Mat:} Tests object-state tracking and intermediate-goal constraints. Three blue mats each hold one block, with a green intermediate mat. The policy must move each block to the green mat and then return it to the center of its original mat. Success requires every block to visit the green mat and finish on its original mat with the robot holding no object at the end.

\end{itemize}

\noindent\textbf{RoboVoLo.}
RoboVoLo is a tabletop manipulation benchmark released with VoLo. It uses Isaac Lab and Isaac Sim for physical simulation, rendering, camera observations, and rigid-body interaction. Tasks pair natural-language instructions with interactive everyday objects and automatic success predicates, evaluating visual understanding, language grounding, and multi-step planning. Its four capability groups are Memory, which requires retaining initial object or container relations during rearrangement; Common Sense, which requires category-based sorting and inference from scene examples; World Knowledge, which combines manipulation with elementary symbolic or world knowledge; and Complex References, which tests compositional references involving negation, ordinals, spatial relations, and size.

Because full-suite evaluation is computationally expensive, we randomly select 10 trials from each task type, yielding 40 evaluation trials. Every selected trial is evaluated with its native progress. The selected tasks are listed below.

\noindent\textbf{Memory.}
\begin{itemize}\setlength{\itemsep}{0pt}\setlength{\parskip}{0pt}\setlength{\parsep}{0pt}
    \item \texttt{RecallBottomToBowlTask}: Unstack two blocks and place the originally bottom block into a bowl.
    \item \texttt{RecallRestackCansTask}: Disassemble a three-can tower and rebuild it in the original order at another table location.
    \item \texttt{RecallStackTopAmongFourTask}: Remember the top block of an initial stack among four blocks and place it into a bowl.
    \item \texttt{RecallTopHalfToBinTask}: Unstack a four-block tower and place the original top two blocks into a bin.
    \item \texttt{SpatialByStackOrderTask}: Place the former bottom object left of the former top object using the initial stack order.
    \item \texttt{SwapBinReplaceBlockCanTask}: Remove a green block from a bin and replace it with a corn can.
    \item \texttt{SwapBinReplaceFruitsTask}: Remove a banana from a container and replace it with an apple.
    \item \texttt{SwapBowlFruitsTask}: Swap a lemon and a lychee between two bowls.
    \item \texttt{ThreeContainerRestoreTask}: Empty three containers, then place the cube and banana initially on the table into the designated bowl.
    \item \texttt{UnstackSelectTask}: Unstack a three-can tower and place the originally top and bottom cans into a bowl.
\end{itemize}

\noindent\textbf{Common Sense.}
\begin{itemize}\setlength{\itemsep}{0pt}\setlength{\parskip}{0pt}\setlength{\parsep}{0pt}
    \item \texttt{InferClearTableTask}: Infer categories from examples and place an orange in the fruit bowl and a cube in the toy box.
    \item \texttt{InferJoinGroupTask}: Infer that a lime belongs with the fruits already in a bowl.
    \item \texttt{InferMajorityGroupTask}: Identify the majority category on the table and place the three fruits into a bowl.
    \item \texttt{InferTripleSortTask}: Sort a lime, crab toy, and corn can using examples in three containers.
    \item \texttt{KitShelfRestockTask}: Balance the number of bottles on the two shelves.
    \item \texttt{KitThreeBinsBlocksTask}: Place a green block in the empty bin so that each of three bins contains one block.
    \item \texttt{RecoverSpillTask}: Recover scattered fruits and cans by sorting them into a wooden bowl and a bin.
    \item \texttt{SortCondimentsVsCansTask}: Sort two condiments and a luncheon-meat can into their corresponding containers.
    \item \texttt{SortFoodVsNonFoodTask}: Sort a lime and a blue block into food and non-food containers.
    \item \texttt{SortFreshVsCannedTask}: Sort a lime and a luncheon-meat can into fresh-food and canned-food containers.
\end{itemize}

\noindent\textbf{World Knowledge.}
\begin{itemize}\setlength{\itemsep}{0pt}\setlength{\parskip}{0pt}\setlength{\parsep}{0pt}
    \item \texttt{ChemCO2Task}: Add two oxygen blocks to a basket containing carbon to form CO$_2$.
    \item \texttt{ChemNobleGasTask}: Select helium from candidate elements and place it in a bin.
    \item \texttt{ChemSaltTask}: Add a chlorine block to a bowl containing sodium to form NaCl.
    \item \texttt{ChemWaterTask}: Add an oxygen block to a bowl containing two hydrogen blocks to form H$_2$O.
    \item \texttt{MathAdditionTask}: Solve $3+4$ and place the number 7 in the answer slot.
    \item \texttt{MathMultiplyTask}: Solve $4\times2$ and place the number 8 in the answer slot.
    \item \texttt{RecycleBananaTask}: Place a banana in the compost bin.
    \item \texttt{RecycleCanTask}: Place an aluminum soda can in the recycling bin.
    \item \texttt{RecycleCupTask}: Place a disposable coffee cup in the general-waste bin.
    \item \texttt{RecycleSort4Task}: Sort a banana, eggplant, soda can, and coffee cup into compost, recycling, and general waste.
\end{itemize}

\noindent\textbf{Complex References.}
\begin{itemize}\setlength{\itemsep}{0pt}\setlength{\parskip}{0pt}\setlength{\parsep}{0pt}
    \item \texttt{XrBiggestFruitTask}: Select the largest orange among several fruits and place it in a bowl.
    \item \texttt{XrExceptOrangeTask}: Place the lemon and lime in a bowl while leaving the orange out.
    \item \texttt{XrExceptTallestTask}: Place every can except the tallest tomato-soup can into a bin.
    \item \texttt{XrLeftOfBlockTask}: Place the three fruits to the left of a red block into a bowl.
    \item \texttt{XrLeftmostRightmostTask}: Place the leftmost lime in a bowl and the rightmost lemon in a bin.
    \item \texttt{XrLeftRightMugTask}: Place the lemon right of a mug in a bowl and the orange left of the mug in a bin.
    \item \texttt{XrNonFruitTask}: Identify the non-fruit can among fruits and place it in a vase.
    \item \texttt{XrSecondFromLeftFiveTask}: Select the second object from the left, a luncheon-meat can, among five objects and place it in a basket.
    \item \texttt{XrSmallestItemTask}: Select the smallest lychee among several objects and place it in a bowl.
    \item \texttt{XrTwoExceptionsTask}: Place all but the red block and tuna can into a bowl.
\end{itemize}

\subsubsection{Real-Robot Protocol}
\label{app:robot}
\begin{figure*}[t]
    \centering
    \includegraphics[width=\textwidth,height=0.8\textheight,keepaspectratio]{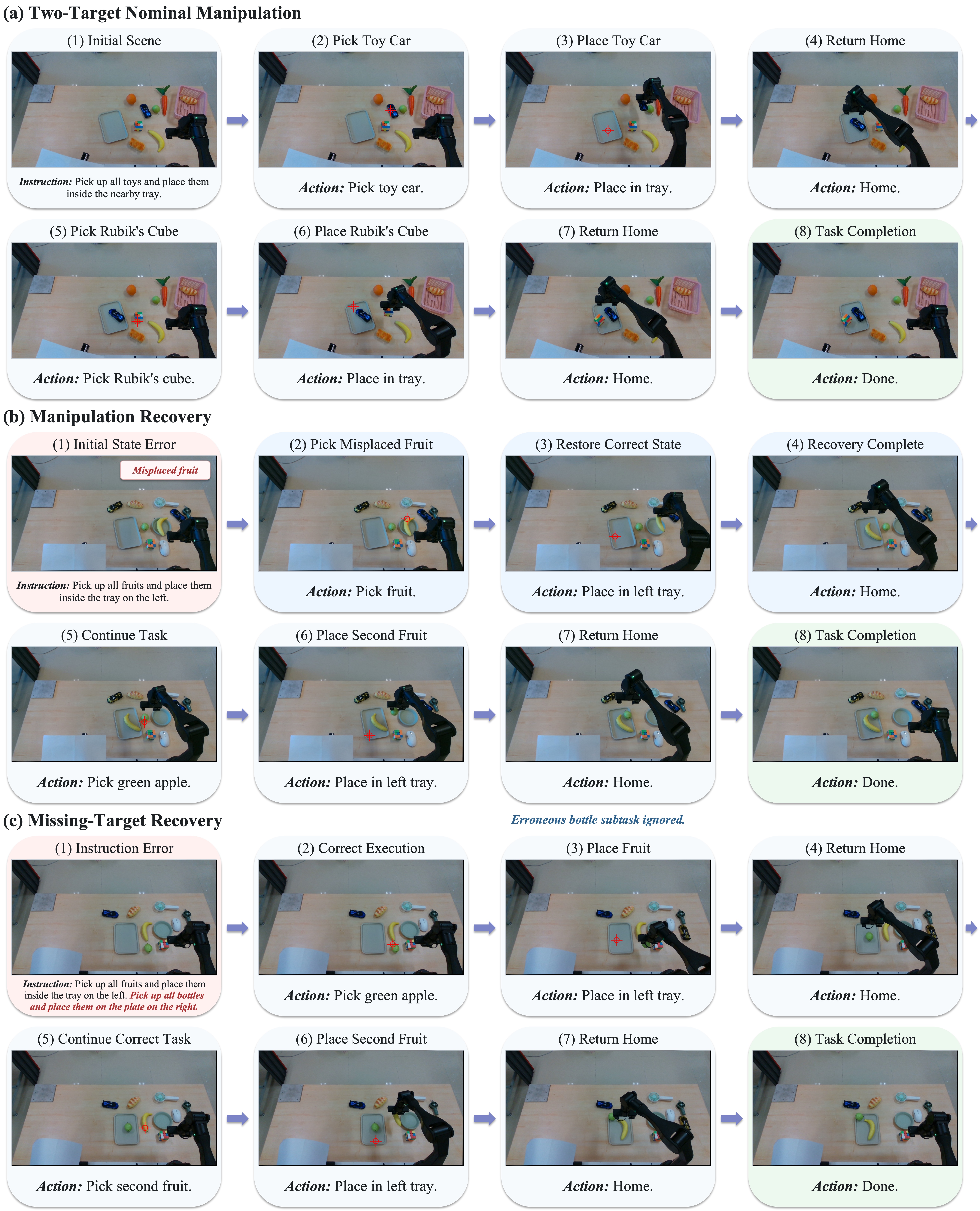}
    \caption{Successful real-robot trajectories. (a) Two-target nominal manipulation. (b) Manipulation Recovery corrects a misplaced fruit before continuing the task. (c) Missing-Target Recovery completes feasible fruit placement while ignoring the unavailable bottle subtask.}
    \label{fig:real-robot-success}
\end{figure*}

\begin{figure*}[t]
    \centering
    \includegraphics[width=\textwidth,height=0.8\textheight,keepaspectratio]{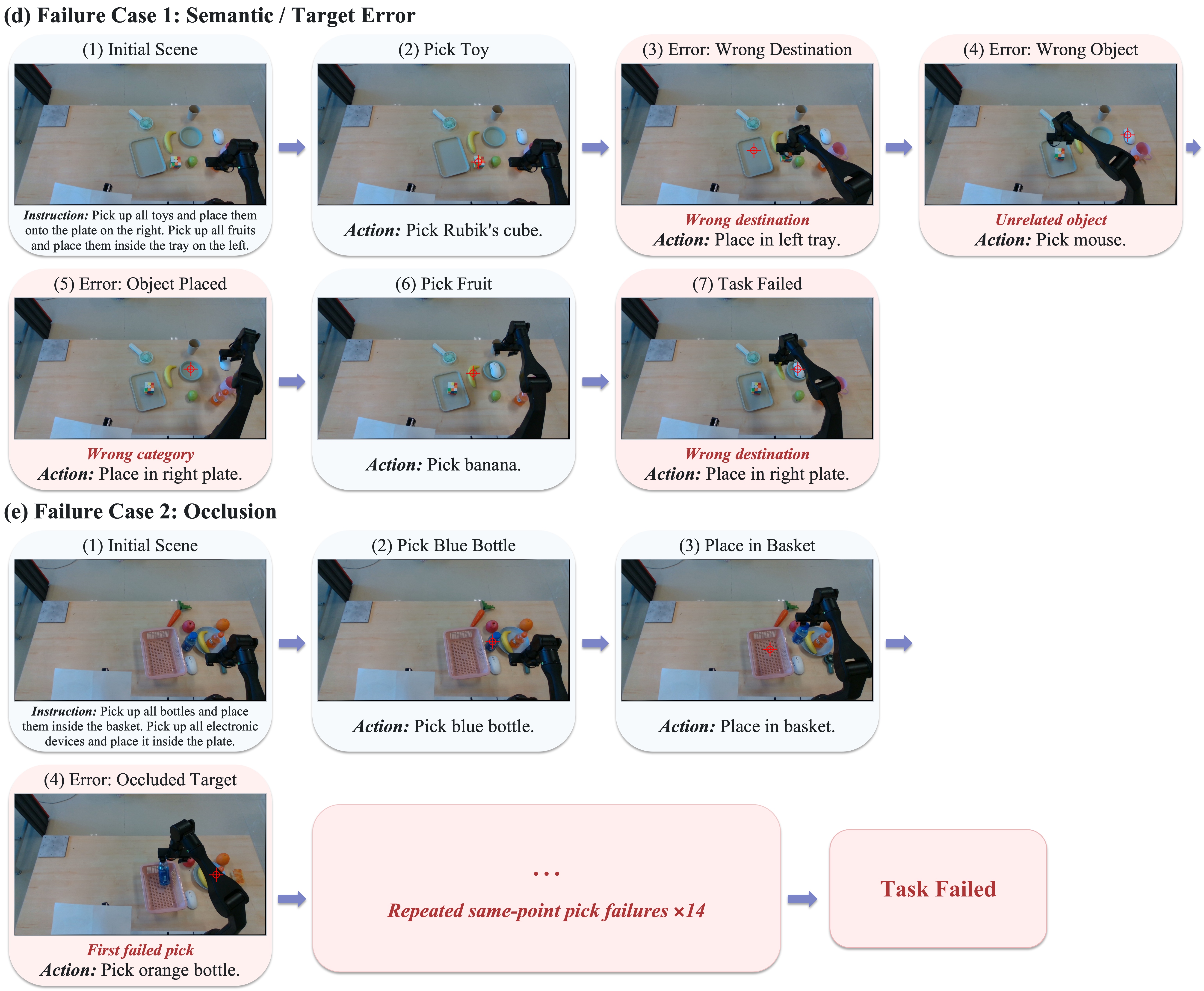}
    \caption{Failure cases in real-robot execution. (d) Semantic or target errors lead to incorrect placements and task failure. (e) An occluded target leads to repeated same-point pick attempts and eventual failure.}
    \label{fig:real-robot-failure}
\end{figure*}

This section describes the real-robot evaluation protocol. We first introduce the hardware setup and evaluation tasks, then describe the policy interface and 2D-to-3D grounding procedure, followed by the control, safety, and success criteria used in real-robot execution.

\noindent\textbf{Hardware and Evaluation Protocol.}
We evaluate three policies on a classic six-axis AIRBOT Play manipulator (AIRBOT Play-01-100) with its end-effector gripper. A fixed Intel RealSense D435i RGB-D camera is mounted above the workspace in an eye-to-hand configuration. We use its aligned color and depth streams only, and do not use inertial measurements. All policies share the same camera calibration, semantic action interface, and low-level control parameters. Each policy is evaluated in 10 independent trials on each of five tasks, for 150 real-robot trials in total. Before each trial, an operator resets the scene and varies object positions and orientations.

\noindent\textbf{Tasks.}
The first three tasks measure nominal long-horizon manipulation, and the remaining two measure recovery:
\begin{itemize}\setlength{\itemsep}{0pt}\setlength{\parskip}{0pt}\setlength{\parsep}{0pt}
    \item \textbf{Two-Target Nominal Manipulation:} Place two toys in a nearby tray. The action budget is 10.
    \item \textbf{Three-Target Nominal Manipulation:} Place toys on a right-side plate and fruits in a left-side tray. The action budget is 15.
    \item \textbf{Four-Target Nominal Manipulation:} Place bottles in a left-side basket and electronic devices on a right-side plate. The action budget is 20.
    \item \textbf{Manipulation Recovery:} Start with a fruit misplaced on a plate and place fruits in the left-side tray. The action budget is 10.
    \item \textbf{Missing-Target Recovery:} Place fruits in the left-side tray and bottles on the right-side plate when no bottle is present. The action budget is 10.
\end{itemize}

Figure~\ref{fig:real-robot-success} and Figure~\ref{fig:real-robot-failure} show representative successful trajectories and failure cases from the real-robot evaluation, respectively.

\noindent\textbf{Policy interface and 2D-to-3D Grounding.}
At each decision step, the policy receives the current RGB observation, task instruction, and ordinary execution result from the preceding action. It selects one of four semantic actions, \texttt{pick}, \texttt{place}, \texttt{home}, or \texttt{done}; \texttt{pick} and \texttt{place} additionally output a normalized image coordinate in $[0,1000]$. The policy does not directly command joint positions, velocities, or trajectories, and it never receives task-specific checker feedback. For \texttt{pick} and \texttt{place}, we first convert the predicted normalized coordinate to the corresponding pixel in the original RGB image and use it as a point prompt for SAM 2~\citep{ravi2024sam2} to obtain a binary mask of the target region. Using the aligned depth image, we collect all valid depth pixels within the predicted mask and back-project them into 3D. Specifically, for each valid mask pixel $(u,v)$ with depth $Z=D(u,v)$, given camera intrinsics $(f_x,f_y,c_x,c_y)$, its 3D coordinate in the camera frame is computed as
\begin{equation}
X=(u-c_x)Z/f_x,\qquad
Y=(v-c_y)Z/f_y,\qquad
Z=D(u,v).
\end{equation}
The resulting set of 3D points forms a target point cloud, which is transformed from the camera frame to the robot base frame using a fixed camera-to-base extrinsic calibration. For \texttt{pick}, we estimate the target surface height using the 95th percentile of the valid point-cloud heights, which is used to determine the grasp height. For \texttt{place}, we identify the valid depth pixel closest to the center of the target mask bounding box and back-project it into 3D to determine the placement location. This grounding pipeline converts the policy's image-space points into geometric targets for the low-level robot controller while keeping the policy interface independent of robot-specific joint control.

\noindent\textbf{Control, Safety, and Success.}
Before execution, the controller verifies that the target and its approach point lie in a predefined Cartesian workspace. A \texttt{pick} opens the gripper, moves to an approach pose above the target, descends, closes the gripper, and retracts. A \texttt{place} moves to an approach pose above the destination, descends, opens the gripper, and retracts.
\texttt{home} moves to a predefined pose without changing gripper state; \texttt{done} ends the task without commanding motion. After each action, the system acquires a new RGB-D observation and returns the ordinary execution result to the policy. Operators check the camera view and workspace before each trial and intervene only to prevent collision, equipment damage, or other safety risks; any intervention counts as a failure. A trial terminates on \texttt{done}, exhaustion of its action budget, an unrecoverable perception or execution error, or safety intervention. We determine success from the final scene state. For Missing-Target Recovery, success requires completing feasible fruit placement, avoiding repeated attempts to grasp the absent bottle, and terminating appropriately.

\noindent\textbf{Evaluation.}
All real-robot evaluation metrics are manually assessed from the recorded trials and final scene states. Task progress, recovery performance, and success are determined from the executed actions and resulting scene configurations. We additionally report Active Re-observation Rate (ARR) to measure whether the policy actively restores visual access when a task-relevant target becomes occluded during execution. For ARR, we manually identify all target-occlusion events across the real-robot trials and count the proportion of these events for which the policy invokes \texttt{home} to restore visual observation. ARR therefore captures a specific recovery behavior: whether the policy recognizes insufficient visual access and actively repositions the robot to reacquire the target before continuing execution. This metric is evaluated independently of overall task success. Specifically, we observe 47 target-occlusion events for the Full ERR-VLM experiments, of which 37 trigger \texttt{home}; 50 events for the SFT model, with no \texttt{home} actions; and 45 events for RoboBrain2.5-8B, of which 4 trigger \texttt{home}.

\end{document}